\documentclass[lettersize,journal]{IEEEtran}
\usepackage{amsmath,amsfonts}
\usepackage{array}
\usepackage[caption=false,font=normalsize,labelfont=sf,textfont=sf]{subfig}
\usepackage{textcomp}
\usepackage{stfloats}
\usepackage{url}
\usepackage{verbatim}
\usepackage{graphicx}
\usepackage{cite}
\usepackage{amsmath}
\usepackage{amsthm}
\newtheorem{definition}{Definition}
\newtheorem{proposition}{Proposition}
\usepackage{algorithm}
\usepackage{algorithmicx}
\usepackage{algpseudocode}
\usepackage{multirow}
\usepackage {epsfig}

\begin{document}

\title{Gradient Aligned Attacks via a Few Queries}

	\author{\IEEEauthorblockN{Xiangyuan Yang\IEEEauthorrefmark{2},  Jie Lin\IEEEauthorrefmark{2}\IEEEauthorrefmark{1}, Hanlin Zhang\IEEEauthorrefmark{3}, Xinyu Yang\IEEEauthorrefmark{2}, and Peng Zhao\IEEEauthorrefmark{2}} \\
	\IEEEauthorblockA{\IEEEauthorrefmark{2} Xi'an Jiaotong University, Xi'an, China,\\
		Emails: ouyang\_xy@stu.xjtu.edu.cn, \{jielin, yxyphd,p.zhao\}@mail.xjtu.edu.cn} \\
	\IEEEauthorblockA{\IEEEauthorrefmark{3} Qingdao University, Qingdao, China, Email:
		hanlin@qdu.edu.cn}
	\thanks{Corresponding author: Jie Lin (jielin@mail.xjtu.edu.cn).}
	\thanks{Manuscript received XXX, XX, 2022; revised XXX, XX, 2022.}}

\markboth{Journal of \LaTeX\ Class Files, ~Vol.~XX, No.~XX, XXX~2022}%
{Shell \MakeLowercase{\textit{Yang et al.}}: Gradient Aligned Attacks via a Few Queries}


\maketitle

\begin{abstract}
Black-box query attacks, which rely only on the output of the victim model, have proven to be effective in attacking deep learning models. However, existing black-box query attacks show low performance in a novel scenario where only a few queries are allowed. To address this issue, we propose gradient aligned attacks (GAA), which use the gradient aligned losses (GAL) we designed on the surrogate model to estimate the accurate gradient to improve the attack performance on the victim model. Specifically, we propose a gradient aligned mechanism to ensure that the derivatives of the loss function with respect to the logit vector have the same weight coefficients between the surrogate and victim models. Using this mechanism, we transform the cross-entropy (CE) loss and margin loss into gradient aligned forms, i.e. the gradient aligned CE or margin losses. These losses not only improve the attack performance of our gradient aligned attacks in the novel scenario but also increase the query efficiency of existing black-box query attacks. Through theoretical and empirical analysis on the ImageNet database, we demonstrate that our gradient aligned mechanism is effective, and that our gradient aligned attacks can improve the attack performance in the novel scenario by 16.1\% and 31.3\% on the $l_2$ and $l_{\infty}$ norms of the box constraint, respectively, compared to four latest transferable prior-based query attacks. Additionally, the gradient aligned losses also significantly reduce the number of queries required in these transferable prior-based query attacks by a maximum factor of 2.9 times. Overall, our proposed gradient aligned attacks and losses show significant improvements in the attack performance and query efficiency of black-box query attacks, particularly in scenarios where only a few queries are allowed.
\end{abstract}


\section{Introduction}
\label{sec:intro}
\IEEEPARstart{D}{eep} neural networks (DNNs)~\cite{VGG,ResNet,Inception-v3,DenseNet,MobileNet} are highly vulnerable to adversarial attacks~\cite{FGSM,I-FGSM,MI-FGSM,ZOO,Square}, particularly black-box query attacks~\cite{ZOO,Square}, which can attack the victim model without any knowledge of its architecture and parameters, making it a serious threat to critical applications such as autonomous driving, finance, and healthcare. This can boost an increased focus on the development of defense methods, such as adversarial training~\cite{TRADES,Fast-AT,Bag-of-tricks-for-AT,Madry-AT} and other methods~\cite{post-process-method}, to protect these applications.

Currently, the black-box query attacks are mainly divided into two categories: the score-based query attacks, which know the probability vector output by the victim model, and the hard label-based query attacks, which only know the predicted label by the victim model. The early proposed score-based query attacks (such as ZOO~\cite{ZOO}, Bandit~\cite{Bandit}, SimBA~\cite{SimBA}, SignHunter~\cite{SignHunter} and NES~\cite{NES}) and latest hard label-based query attacks (including boundary attack~\cite{boundary-attack}, OPT attack~\cite{opt-attack}, hop skip jump attack~\cite{HopSkipJumpAttack} and Sign-OPT~\cite{Sign-opt}) require an average of hundreds of queries to execute a successful attack. However, this becomes expensive when attacking a lot of examples because of \$1.5 for 1000 image queries on Google Cloud Vision API~\cite{SimBA}. To reduce the number of queries, the latest proposed score-based query attacks (such as PRGF~\cite{P-RGF}, TREMBA~\cite{TREMBA}, LeBA~\cite{LeBA}, ODS~\cite{ODS} and GFCS~\cite{GFCS}) used the transferable prior from the surrogate model as the initially estimated gradient, which can reduce a large number of queries because of its good transferability~\cite{MI-FGSM,DI-FGSM,SI-NI-FGSM,VMI-FGSM}. Particularly, the GFCS attack~\cite{GFCS} can reduce the number of queries to less than 10. Therefore, due to the expensive query fees~\cite{SimBA}, query limitation~\cite{NES} and high effectiveness of the GFCS attack~\cite{GFCS}, we consider a novel scenario where only a few queries are allowed (i.e., the maximum number of queries is less than 10). However, in the novel scenario, the latest transferable prior-based query attacks~\cite{P-RGF,LeBA,ODS,GFCS} exhibited low performance in forms of attacking the victim model successfully.

\begin{figure}[t]
\begin{center}
\centerline{\includegraphics[width=\columnwidth]{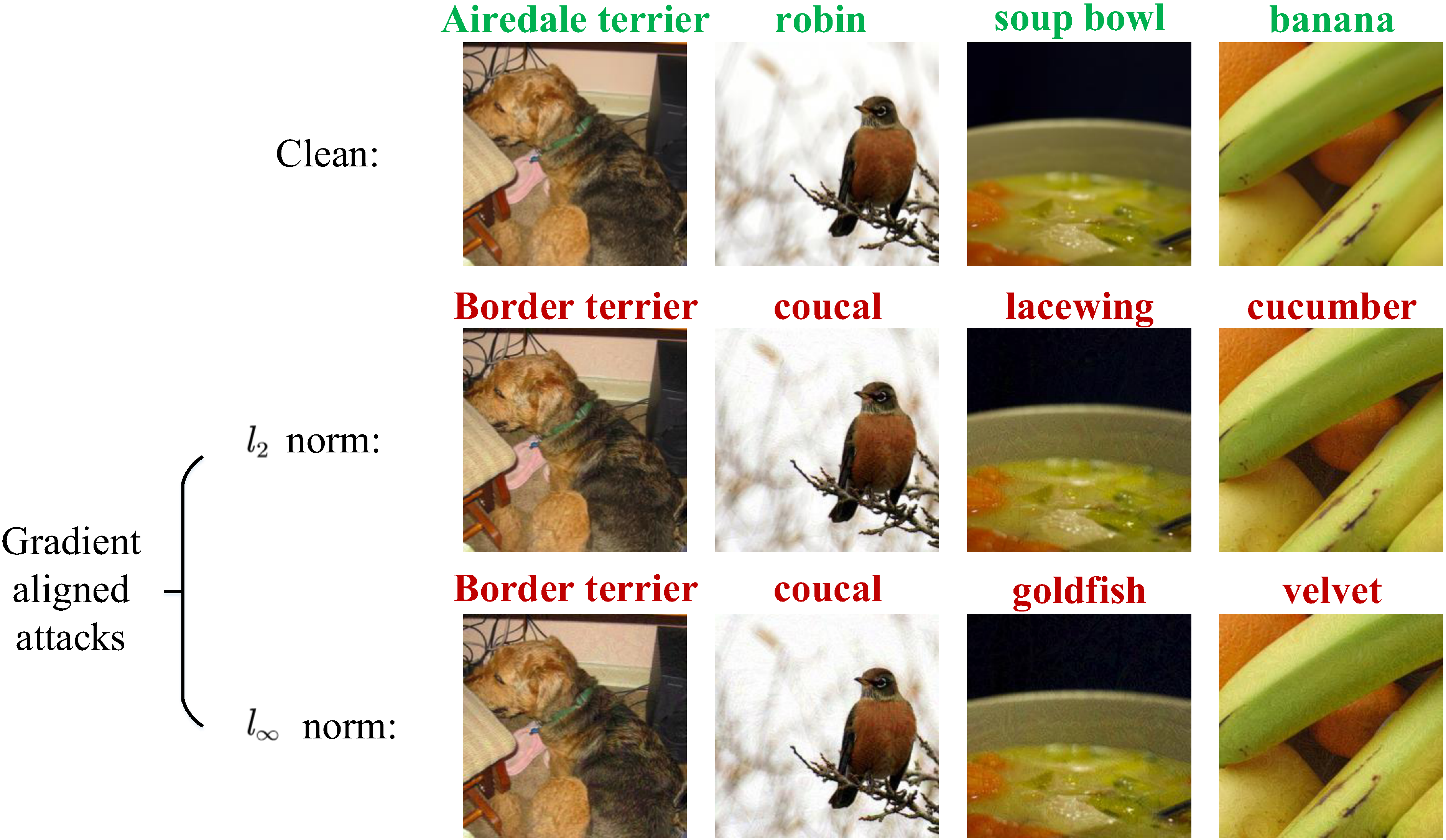}}
\caption{Four clean examples and their corresponding adversarial examples generated using the gradient aligned attacks under the $l_2$ and $l_{\infty}$ norms of the box constraint. The correct category is denoted in green, while the wrong category is denoted in red. We use ResNet50~\cite{ResNet} as the surrogate model and VGG16~\cite{VGG} as the victim model.}\vspace{-5mm}
\label{FIG:adversarial_examples}
\end{center}
\end{figure}

To address the aforementioned issue, we propose gradient aligned attacks that aim to improve the attack success rate (ASR) on the victim model in a novel scenario where only a few queries are allowed. Specifically, we observe the preference property of deep learning models that: the generated adversarial examples are preferred to be misclassified as the wrong category with the highest probability in the process of generating adversarial examples with the untargeted attacks, where the wrong category refers to any category other than the ground truth. Moreover, we draw on the concept of gradient alignment~\cite{gradient-alignment}, which shows that a higher cosine similarity of the gradients between the surrogate and victim models leads to higher transferability. However, the authors of \cite{gradient-alignment} did not attempt to enhance the metric of the gradient alignment. Inspired by the preference property and gradient alignment, we propose a gradient aligned mechanism that can actively enhance the metric of the gradient alignment. The gradient aligned mechanism can change the cross-entropy (CE) loss and margin loss into gradient aligned forms, namely the gradient aligned CE or margin losses. Our proposed gradient aligned attacks utilize these losses to significantly improve the attack success rates of current transfer attacks such as MIFGSM~\cite{MI-FGSM}, NIFGSM~\cite{SI-NI-FGSM}, VMIFGSM~\cite{VMI-FGSM}, among others~\cite{SGM,MoreBayesian}. In the novel scenario, our gradient aligned attacks exhibit higher attack success rates than the latest query attacks~\cite{P-RGF,LeBA,ODS,GFCS} by a large margin. Additionally, the gradient aligned losses can also be used to enhance the query efficiency of the latest transferable prior-based query attacks. Fig.~\ref{FIG:adversarial_examples} shows the resulting adversarial examples generated by our gradient aligned attacks are visually similar to the original examples but cause the DNN classifier to misclassify them as a different category.

Our main contributions are summarized as follows:
\begin{itemize}
\item In black-box query attacks, the query overhead can be quite expensive, and there are also some limitations on the number of queries allowed. Therefore, we consider a novel scenario where only a limited number of queries (i.e., less than 10) are allowed.
\item We have identified a preference property in DNN classifiers where the generated adversarial examples tend to be misclassified as the category with the highest probability, except for the ground truth category, during the process of generating adversarial examples. Leveraging this property, we propose the gradient aligned mechanism to improve the metric of gradient alignment. This mechanism ensures that the derivatives of the loss function with respect to the logit vector have the same weight coefficients for both the surrogate and victim models.
\item By employing the gradient alignment mechanism, we transform the commonly used CE loss and margin loss into their gradient aligned forms, i.e., the gradient aligned CE or margin losses. Then the gradient aligned attacks are proposed to enhance the attack performance of the black-box query attacks using the gradient aligned losses in a novel scenario where only a few queries are allowed.
\item The proposed gradient aligned losses can not only be readily incorporated into the transferable prior-based query attacks to enhance their query efficiency, but also improve the attack success rates of the latest transfer attacks by increasing the metric of the gradient alignment between the surrogate and victim models.
\item Through theoretical and empirical analysis on ImageNet database, we have found that the gradient aligned attacks can significantly improve the attack success rates by 16.1\% and 31.3\% on the $l_2$ and $l_{\infty}$ norms of the box constraint, respectively, compared to the latest transferable prior-based query attacks. Moreover, the proposed gradient aligned losses can significantly reduce the average number of queries required for the latest transferable prior-based query attacks by a factor of 1.2 to 2.9 times.
\end{itemize}

In the remainder of this paper, Section~\ref{sec:preliminary} introduces the main notations used in this paper, common loss functions used in black-box attacks, and the latest basic ideas in the areas of black-box query attacks and transfer attacks. Then, Section~\ref{sec:methodology} proposes the gradient aligned mechanism, losses, and attacks to address the low attack performance of the current query attacks in the proposed novel scenario. Finally, we conduct experiments to demonstrate the effectiveness of the proposed methods in Section~\ref{sec:experiments}, summarize the related work in Section~\ref{sec:related_work}, and conclude this paper in Section~\ref{sec:conclusion}.

\section{Preliminary}
\label{sec:preliminary}
In this section, we first define the notations used in this paper. Then, the commonly used loss functions in the query attacks are introduced. Finally, we provide an overview of the current black-box attacks and their development history.

\begin{table}
	\centering \caption{Notations}\label{tab:notations}
	\begin{tabular}{lp{0.33\textwidth}} \hline
		$f:$& The victim model.\\
		$\hat{f}:$& The surrogate model.\\
		$x:$& The natural example.\\
		$y:$& The ground truth of $x$.\\
		$y_\tau:$& The target label in the targeted attacks.\\
		$x^{adv}:$& The corresponding adversarial example of $x$.\\
		$\epsilon:$& The magnitude of the adversarial perturbations, i.e. the attack strength.\\
		$\alpha:$& The step size in each iteration.\\
		$\delta:$& The perturbation added into $x^{adv}$.\\
		$f(\cdot), \hat{f}(\cdot):$& The logit vectors outputed by $f$ and $\hat{f}$, respectively.\\
		$\boldsymbol{z}_x:$& The logit vector of $x$.\\
		$C:$& The number of categories in the classification task.\\
		\hline
	\end{tabular}
\end{table}

\subsection{Notations}
\label{sec:notations}
The main notations are defined in Table~\ref{tab:notations}. Given a natural example pair $(x,y)$, an adversary attempts to generate an imperceptible adversarial example $x^{adv}$ that is close to the natural example $x$, with the goal of misclassifying it under the victim classifier $f$, i.e.,
\begin{gather}
\left\{ \begin{array}{c}
arg\max f(x^{adv}) \ne y, if\,\,untargeted\,\,attack\\
arg\max f(x^{adv}) = y_{\tau}, if\,\,targeted\,\,attack\\
\end{array} \right.\label{eq:target_of_adversarial_example}
\end{gather}

\begin{align}
x^{adv}=x+\delta, x^{adv}\in\mathcal{B}_p(x,\epsilon) \label{eq:consist_of_adversarial_example}
\end{align}
where $y_\tau$ is the target label in the target attacks, $\delta$ is the added perturbation and the generated adversarial example $x^{adv}$ is constrained to an $l_p$ ball centered at $x$ with radius $\epsilon$. This paper mainly focuses on $l_2$ and $l_\infty$ attacks to align with the previous works~\cite{GFCS,Bandit,P-RGF}.

\subsection{Loss Functions}
\label{sec:loss_functions}
The cross-entropy (CE) loss function is widely used to guide the process of generating adversarial examples. It measures the difference between the predicted probability distribution and the true one-hot label, i.e.,
\begin{align}
L_{CE}(x,y)=-\log [S(f(x))]_y \label{eq:ce_loss}
\end{align}
where $S(\cdot)$ indicates the softmax function and $[\cdot]_y$ denotes the value with index $y$ in a vector.

In addition, the margin loss is also commonly used in adversarial attacks, which aims to maximize the difference between the correct class score and the other classes' scores by a margin, i.e.,
\begin{align}
L_{margin}(x,y)=\max_{i\ne y}([f(x)]_i-[f(x)]_y). \label{eq:margin_loss}
\end{align}

\subsection{Black-box Query Attacks}
\label{sec:black_box_query_attacks}
The black-box query attacks only have access to the output of the victim model without any knowledge of its architecture and parameters, which makes it impossible to directly calculate the gradient by taking the derivative of the loss function with respect to the input. To estimate the gradient, Chen et al.~\cite{ZOO} proposed \textbf{the zeroth-order optimization stochastic gradient descent with coordinate-wise Newton's method (ZOO)}, which can be formulated as follows:
\begin{gather}
g_{t+1}\approx \frac{f(x_{t}^{adv}+\lambda \boldsymbol{e}_i)-f(x_{t}^{adv}-\lambda \boldsymbol{e}_i)}{2\lambda} \label{eq:first_order_difference} \\
h_{t+1}\approx \frac{f(x_{t}^{adv}+\lambda \boldsymbol{e}_i)-2f(x_t^{adv})+f(x_{t}^{adv}-\lambda \boldsymbol{e}_i)}{\lambda^2} \label{eq:second_order_difference} \\
x_{t+1}^{adv}=\left\{ \begin{array}{c}
x_{t}^{adv}+\alpha \cdot g_{t+1}, h_{t+1}\leq 0\\
x_{t}^{adv}+\alpha \cdot \frac{g_{t+1}}{h_{t+1}}, h_{t+1}>0\\
\end{array} \right. \label{eq:zoo_update}
\end{gather}
where $\lambda$ is a small constant, $\boldsymbol{e}_i$ indicates a randomly chosen standard basis vector, $g_{t+1}$, $h_{t+1}$ and $x^{adv}_{t+1}$ respectively denote the estimated gradient, the estimated Hessian matrix and the generated adversarial example at $t+1$-th iteration. Note that, the update using $\frac{g_{t+1}}{h_{t+1}}$ has faster convergence rate than using $g_{t+1}$ when $h_{t+1}$ is greater than 0.

The derivative-free of the zeroth-order optimization method, \textbf{NES}~\cite{NES}, was proposed to improve the query efficiency of the query attacks by maximizing the expected value of the loss function under the search distribution, i.e.,
\begin{gather}
g_{t+1}=\sum_i^n{[S(f(x_{t}^{adv}+\sigma \cdot u_i))]_y\cdot u_i}\nonumber\\
-\sum_i^n{[S(f(x_{t}^{adv}-\sigma\cdot u_i))]_y\cdot u_i} \label{eq:nes_gradient}\\
u_i\sim \mathcal{N} (\mathbf{0}_N,\boldsymbol{I}_{N\times N}) \\
x_{t+1}^{adv}=x_{t}^{adv}+\alpha \cdot \frac{g_{t+1}}{2n\sigma} \label{eq:nes_update}
\end{gather}
where $\mathcal{N} (\mathbf{0}_N,\boldsymbol{I}_{N\times N})$ is a multivariate normal distribution with a mean of $\mathbf{0}_N$ and a covariance of $\boldsymbol{I}_{N\times N}$, $N$ indicates the dimension of the input, $n$ is the number of samples and $\sigma$ represents the search variance.

Guo et al.~\cite{SimBA} demonstrated that \textbf{A simple black-box attack (SimBA)} can significantly reduce the number of queries by directly using the basis vector as the estimated gradient, without decreasing the attack success rates, i.e.,
\begin{gather}
x_{t+1}^{adv}=\left\{ \begin{array}{c}
x_{t}^{adv}+\alpha \cdot \boldsymbol{q}, if\,\,[S(f(x_{t}^{adv}))]_y\\
>[S(f(x_{t}^{adv}+\alpha\cdot\boldsymbol{q}))]_y\\
x_{t}^{adv}-\alpha\cdot\boldsymbol{q}, else\,\,if\,\,[S(f(x_{t}^{adv}))]_y\\
>[S(f(x_{t}^{adv}-\alpha\cdot\boldsymbol{q}))]_y\\
\end{array}\right. \label{eq:simba_update}\\
\boldsymbol{q}\in Q 
\end{gather}
where $Q$ is a set of basis vectors.

\textbf{Bandits-TD~\cite{Bandit}} proposed a different approach to estimating gradients. The authors argue that only a fraction of the true gradient direction is sufficient for a successful attack, and therefore, they introduce the notion of the average-blurred gradient. This involves averaging the gradient across neighboring pixels, reducing the dimension of the estimated gradient and improving the query efficiency of the attack.

Based on the argument in Bandits-TD~\cite{Bandit}, \textbf{SignHunter~\cite{SignHunter}} transformed continuous black-box optimization into binary optimization. Instead of relying on the randomness of the estimated gradient $\boldsymbol{q}$ in previous methods, SignHunter attempts to determine the gradient using past queries. Moreover, SignHunter transforms an $N$-dimensional binary optimization into $N$ one-dimensional binary optimizations, which reduces the query space from $2^{N}$ to $2\times N$, i.e., 
\begin{gather}
\underset{\boldsymbol{q}\in\mathcal{H}}{\max}\:\mathcal{D} _{\boldsymbol{q}}L_{CE}(x,y)=\underset{\boldsymbol{q}\in \mathcal{H}}{\max}\:\boldsymbol{q}^{\top}\cdot\nabla _xL_{CE}(x,y)\Rightarrow \nonumber\\
\underset{\boldsymbol{q}\in\mathcal{H}}{\max}\:\mathcal{D} _{\boldsymbol{q}}L_{CE}(x,y)=\sum_i^N{\underset{q_i\in\{-1,1\}}{\max}q_i[\nabla_xL_{CE}(x,y)]_i} \label{eq:signhunter_update} \\
\mathcal{H} \equiv \{-1,1\}^N 
\end{gather}
where $\boldsymbol{q}$ is a binary gradient.

\textbf{Square~\cite{Square}} analyzed the properties of adversarial examples generated by $l_2$ and $l_\infty$ attacks. They found that the perturbation of the $l_\infty$-norm of adversarial examples in each part is basically $\pm\epsilon$, except for some parts clipped in $[0,1]$, while the perturbation of the $l_2$-norm of adversarial examples concentrates on the local region. Therefore, Square proposed two sampling distribution algorithms for $l_2$ and $l_\infty$ attacks, respectively, to improve the attack success rates under a limited number of queries.

Recently, to further improve query efficiency, several methods have been proposed, including TREMBA~\cite{TREMBA}, PRGF~\cite{P-RGF}, ODS~\cite{ODS}, LeBA~\cite{LeBA}, and GFCS~\cite{GFCS}. These methods use the transferable prior of a surrogate model to initialize the estimated gradient. Specifically, \textbf{ODS~\cite{ODS}} attempts to maximize the output diversity of the surrogate model to improve the transferability of the estimated gradient, i.e., 
\begin{align}
g=\frac{\nabla_x(\boldsymbol{\omega}_{d}^{\top}\cdot \hat{f}(x))}{\left\|\nabla_x(\boldsymbol{\omega}_{d}^{\top}\cdot \hat{f}(x))\right\|_2} \label{eq:ods_gradient}
\end{align}
where the vector $\boldsymbol{\omega}_{d}$ is sampled from the uniform distribution $[-1,1]^C$ and $C$ is the number of categories in the classification task. \textbf{LeBA~\cite{LeBA}} utilizes the surrogate model not only to estimate the gradient but also to update its parameters using feedback query results, improving the adaptability of the surrogate model transferred to the victim model.

When an adversary is not allowed to query the score of each category in the victim model but only achieves the predicted label, the above score-based query attacks become ineffective. To address this issue, several hard label-based query attacks have been proposed, including boundary attack~\cite{boundary-attack}, OPT attack~\cite{opt-attack}, hop skip jump attack~\cite{HopSkipJumpAttack}, and Sign-OPT~\cite{Sign-opt}. Among them, \textbf{boundary attack~\cite{boundary-attack}} and \textbf{hop skip jump attack~\cite{HopSkipJumpAttack}} minimize the perturbation from an initialized adversarial example with large disturbance. On the other hand, \textbf{OPT attack~\cite{opt-attack}} and \textbf{Sign-OPT~\cite{Sign-opt}} transform the hard label-based query attacks into continuous real-value optimization, i.e.,
\begin{align}
\min \mathcal{D}(\boldsymbol{\theta})=\min \underset{\lambda>0}{arg\min}(f(x+\lambda\frac{\boldsymbol{\theta}}{\left\| \boldsymbol{\theta} \right\|})\ne y)
\label{eq:opt_attack}
\end{align}
where $\boldsymbol{\theta}$ represents the search direction and $\lambda$ denotes the minimum distance from the natural input $x$ to its nearest adversarial example along the direction $\boldsymbol{\theta}$.

\subsection{Black-box Transfer Attacks}
\label{sec:black_box_transfer_attacks}
The black-box transfer attacks do not require any knowledge of the victim model, including its output. These types of attacks assume that there is transferability between models with similar tasks. Therefore, the goal of this type of attack is to improve the transferability of the adversarial examples generated on the surrogate model. The \textbf{Fast Gradient Sign Method (FGSM)~\cite{FGSM}} was the first proposed transfer attack, which can be formulated as follows:
\begin{align}
x^{adv}=x+\epsilon\cdot sign(\nabla_xL_{CE}(\hat{f}(x),y))
\label{eq:fgsm_update}
\end{align}
where $sign(\cdot)$ represents the sign function.

\textbf{Iterative FGSM (IFGSM)~\cite{I-FGSM}} updated the adversarial examples with multiple small steps, i.e.,
\begin{gather}
x_{t+1}^{adv}=x_t^{adv}+\alpha\cdot sign(\nabla_{x_{t}^{adv}}L_{CE}(\hat{f}(x_{t}^{adv}),y)) \label{eq:ifgsm_update}\\
x_0^{adv}=x\\
x^{adv}=x_T^{adv}
\end{gather}
$T$ is the number of steps and the step size is $\alpha$.

\textbf{Momentum-based IFGSM (MIFGSM)~\cite{MI-FGSM}} added a momentum term to each update iteration of IFGSM~\cite{I-FGSM} to robustly determine the update direction. The update equation of MIFGSM can be formulated as follows:
\begin{gather}
x_{t+1}^{adv}=x_t^{adv}+\alpha\cdot sign(g_{t+1}) \label{eq:mifgsm_update}\\
g_{t+1}=\mu\cdot g_t+\frac{\nabla_{x_t^{adv}}L_{CE}(\hat{f}(x_t^{adv}),y)}{\left\|\nabla_{x_{t}^{adv}}L_{CE}(\hat{f}(x_t^{adv}),y)\right\|_1} \label{eq:mifgsm_momentum}
\end{gather}
where $\mu$ is the decay factor of the momentum.

\textbf{Nesterov accelerated method~\cite{SI-NI-FGSM}, NIFGSM} is an improvement over MIFGSM~\cite{MI-FGSM} that enhances the convergence rate. In this method, $\hat{f}(x_t^{adv})$ in Equation~\ref{eq:mifgsm_momentum} is replaced with $\hat{f}(x_t^{adv}+\alpha\cdot\mu\cdot g_t)$.

\textbf{Variance tuning-based MIFGSM (VMIFGSM)~\cite{VMI-FGSM}} further stabilize the udpate direction of MIFGSM~\cite{MI-FGSM} using the variance tuning method. In this method, the variance $v_{t+1}$ is calculated by Equation~\ref{eq:vmifgsm_variance} to tune the gradient $g_{t+1}$ in Equation~\ref{eq:mifgsm_momentum}:
\begin{gather}
v_{t+1}=\frac{1}{M}\cdot\sum^M_{i=1}\nabla_{x_{ti}^{adv}}L_{CE}(\hat{f}(x_{ti}^{adv}),y)-\nonumber \\
\nabla_{x_t^{adv}}L_{CE}(\hat{f}(x_t^{adv}),y) \label{eq:vmifgsm_variance}\\
x_{ti}^{adv}=x_t^{adv}+r_i, r_i\sim [-(\beta\cdot \epsilon),(\beta\cdot \epsilon)]^N
\end{gather}
where $[-(\beta\cdot \epsilon),(\beta\cdot \epsilon)]^N$ is a $N$-dimensional uniform distribution and $\beta$ is a hyperparameter.

\section{Methodology}
\label{sec:methodology}
In this section, we present a novel scenario for black-box attacks and propose a solution to adapt to this scenario. First, we investigate the properties of deep learning models and propose the gradient-aligned mechanism. Using this mechanism, we introduce the gradient-aligned losses that are derived from commonly used losses, such as the cross-entropy loss and margin loss. Finally, we utilize these losses to construct the gradient-aligned attacks, which achieve the best results in the novel scenario.

\subsection{A Novel Scenario}
\label{sec:a_novel_scenario}

\begin{table}[t]
\caption{The untargeted attack success rates of different pure query attacks on VGG16~\cite{VGG} in the novel scenario.}
\renewcommand\arraystretch{1.0}
\footnotesize
\centering
\resizebox{1.0\columnwidth}{!}{
\begin{tabular}{ccc|ccc}
\hline
\multicolumn{3}{c|}{$l_\infty$ attacks} & \multicolumn{3}{c}{$l_2$ attacks} \\ \hline
SignHunter  & Bandit-TD  & Square & SimBA   & Bandit-TD  & Square  \\
3.4\%       & 5.7\%      & 38.2\% & 0.5\%   & 6.6\%      & 22.9\%  \\ \hline
\end{tabular}
}
\label{tab:pure_query_attack_in_the_novel_scenario}
\end{table}

Guo et al.~\cite{SimBA} claimed that conducting 1000 image queries would cost \$1.5 on Google Cloud Vision API. However, evaluating the performance of current query attacks on real systems can be expensive due to the large number of queries required, which could hinder the development of trustworthy artificial intelligence. Additionally, the query limitation~\cite{NES} significantly reduces the performance of most query attacks~\cite{ZOO,Bandit,SimBA,Square,SignHunter,NES}. In this paper, we propose a novel scenario in which only a few queries (i.e., less than 10) are permitted. The proposed algorithm for this scenario is minimally affected by query limitations and can be effectively used to evaluate the robustness of defense methods such as adversarial training~\cite{TRADES,Dual-Path-Distillation,Fast-AT,Bag-of-tricks-for-AT,Madry-AT}.


In the novel scenario, the current hard label-based query attacks~\cite{boundary-attack,opt-attack,HopSkipJumpAttack,Sign-opt} are likely to fail due to two main reasons. Firstly, computing the initialized adversarial examples~\cite{boundary-attack,HopSkipJumpAttack} can require several queries, which has consumed the query budget. Secondly, the gradient estimation in each iteration may cost more than 10 queries~\cite{opt-attack,Sign-opt}, which exceeds the query budget. As shown in Table~\ref{tab:pure_query_attack_in_the_novel_scenario}, the pure score-based query attacks~\cite{Bandit,SimBA,Square,SignHunter,NES} that do not use the transferable prior of the surrogate model show low attack success rates in the novel scenario, especially Bandit-TD~\cite{Bandit}, SignHunter~\cite{SignHunter}, and SimBA~\cite{SimBA}. Although transferable prior-based query attacks~\cite{TREMBA,P-RGF,ODS,LeBA,GFCS} have great query efficiency, they still cannot adapt to the novel scenario as shown in Section~\ref{sec:GAA_in_the_novel_scenario}. Therefore, these issues have motivated us to design a novel query attack algorithm that is adapted to the proposed novel scenario.

\subsection{Gradient Aligned Mechanism}
\label{sec:gradient_aligned_mechanism}
Inspired by the concept of gradient alignment~\cite{gradient-alignment}, where a high transfer attack success rate is achieved when the cosine similarity between the gradients of the surrogate and victim models is large, we first explore the preference property of deep learning models in Proposition~\ref{proposition1}. This property can help improve the metric of gradient alignment. Before introducing Proposition~\ref{proposition1}, we define the terms "top-$\bar{n}$ wrong categories" and "top-$\bar{n}$ wrong categories attack success rate" (top-$\bar{n}$ ASR) in Definition~\ref{definition1}.

\begin{definition}
\textbf{(Top-$\bar{n}$ wrong categories and top-$\bar{n}$ wrong categories attack success rate (top-$\bar{n}$ ASR))}
Given an example $(x,y)$, the logits output of the victim model is denoted as $f(x)$. The top-$\bar{n}$ wrong categories are $\bar{n}$ categories with the largest value in $f(x)$ except for the ground truth $y$. They are denoted as $\{y_{\tau_i}|i\leq \bar{n}\}$. The top-$\bar{n}$ ASR is defined as the accuracy of the generated adversarial example $x^{adv}$ classified as a wrong category in the top-$\bar{n}$ wrong categories.
\label{definition1}
\end{definition}

\begin{proposition}
When the victim model is attacked by the gradient-based iterative attacks~\cite{I-FGSM} under the white-box setting, the successfully attacked adversarial examples prefer to be classified as a wrong category in the top-$\bar{n}$ wrong categories, i.e.,
\begin{align}
\label{eq:preference_property}
arg\max f(x^{adv})\in\left\{ y_{\tau_i}|i\leq \bar{n}\right\}
\end{align}
\label{proposition1}
\end{proposition}

\begin{figure*}[t]
\begin{center}
\centerline{\includegraphics[width=\textwidth]{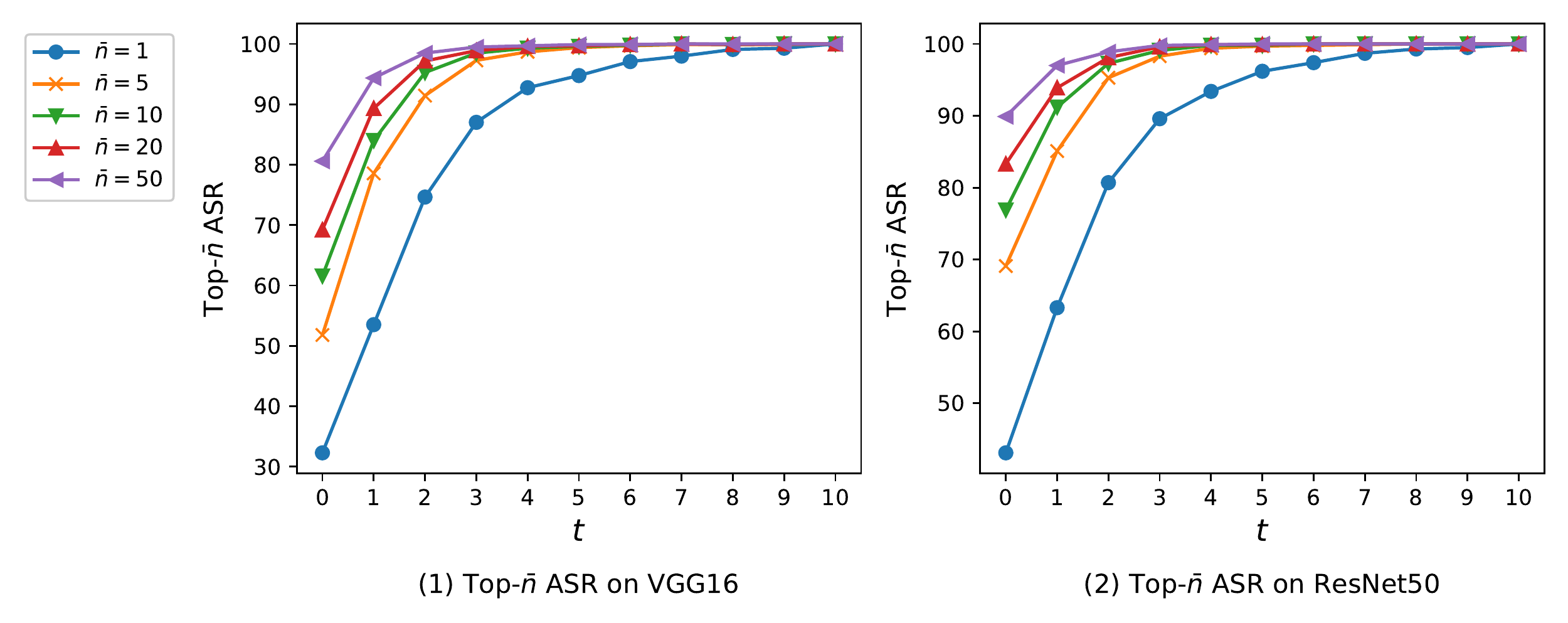}}
\caption{The top-$\bar{n}$ ASR (\%) curves on VGG16~\cite{VGG} and ResNet50~\cite{ResNet} under the white-box setting when varying the iteration $t$.}\vspace{-5mm}
\label{FIG:proposition1}
\end{center}
\end{figure*}

\begin{proof}[Empirical Proof]
As shown in Fig.~\ref{FIG:proposition1}, we plot the top-$\bar{n}$ ASR curves on VGG16~\cite{VGG} and ResNet50~\cite{ResNet} under the white-box setting where the adversarial examples are generated by IFGSM~\cite{I-FGSM} and the top-$\bar{n}$ ASR is calculated by Equation~\ref{eq:top_n_ASR}.
\begin{align}
\label{eq:top_n_ASR}
ASR_t=\frac{\left| \left\{ x_{t}^{adv}|arg\max f(x_{t}^{adv})\in\left\{ y_{\tau_i}^{t-1}|i\leq \bar{n}\right\} \right\} \right|}{\left| \mathcal{S} \right|}
\end{align}
where $ASR_t$ is the top-$\bar{n}$ ASR at $t$-th iteration, $\{y_{\tau_i}^{t-1}|i\leq\bar{n}\}$ is the top-$\bar{n}$ wrong categories of $x_{t-1}^{adv}$ and the test set is $\mathcal{S}$. The results in Fig.~\ref{FIG:proposition1} demonstrate that i) the top-$\bar{n}$ ASR approaches 100\% as the iteration $t$ increases, and ii) increasing the size of the parameter $\bar{n}$ slightly can significantly improve the top-$\bar{n}$ ASR. Based on these empirical observations, we can conclude that the proposition stating that the successful adversarial examples generated by gradient-based iterative attacks tend to be classified as a wrong category in the top-$\bar{n}$ wrong categories under the white-box setting is correct.
\end{proof}

\begin{proof}[Theoretical Proof]
Assuming that the CE loss in Equation~\ref{eq:ce_loss} is used in the gradient-based iterative attacks under the white-box setting. The derivation formula of $L_{CE}$ with respect to the input $x$ is
\begin{gather}
\frac{\partial L_{CE}}{\partial x}=\frac{\partial L_{CE}}{\partial z_y}\cdot \frac{\partial z_y}{\partial x}+\sum_{i=1( i\ne y )}^C{\frac{\partial L_{CE}}{\partial z_i}\cdot \frac{\partial z_i}{\partial x}}=-\frac{1}{\ln 2}\cdot \nonumber\\
( 1-\frac{e^{z_y}}{\sum\nolimits_{i=1}^C{e^{z_i}}} ) \cdot \frac{\partial z_y}{\partial x}+\frac{1}{\ln 2}\cdot \sum_{i=1( i\ne y )}^C{\frac{e^{z_i}}{\sum\nolimits_{j=1}^C{e^{z_j}}}}\cdot \frac{\partial z_i}{\partial x} \nonumber \\
=-\frac{1}{\ln 2}\cdot(1-p_y)\cdot\frac{\partial z_y}{\partial x}+\frac{1}{\ln 2}\cdot\sum_{i=1( i\ne y )}^C{p_i}\cdot \frac{\partial z_i}{\partial x}
\label{eq:derivation_formula_ce}\\
z_i=[f(x)]_i \label{eq:logit} \\
p_i=\frac{e^{z_i}}{\sum\nolimits_{j=1}^C{e^{z_j}}} \label{eq:probability}
\end{gather}
where $z_i$ is the logit of the category with index $i$ and its probability is $p_i$. In Equation~\ref{eq:derivation_formula_ce}, the coefficient of $\frac{\partial z_i}{\partial x}$ is $\frac{p_i}{\ln 2}$ for any wrong category $i$. As the probability $p_i$ of a wrong category increases, so does the coefficient, which means that successful adversarial examples tend to be misclassified as the category with the highest probability among the wrong categories. The proof of the margin loss function is similar.
\end{proof}

Leveraging the preference property in Proposition~\ref{proposition1}, we propose the gradient aligned mechanism in Equation~\ref{eq:gradient_aligned_mechanism}, which ensures that the derivative of the loss function with respect to the logit vector between the surrogate and victim models has the same weight coefficients. Therefore, the loss function used on the surrogate model is changed to:
\begin{gather}
L_{GA}(x,y;\hat{f})=\sum_i^C{\frac{\partial L(x,y;f)}{\partial z_i^f}\cdot z_i^{\hat{f}}} \label{eq:gradient_aligned_mechanism}\\
z_i^f = [f(x)]_i\\
z_i^{\hat{f}} = [\hat{f}(x)]_i
\end{gather}
where $L(x,y;f)$ is the loss function with respect to the victim model $f$ and $L_{GA}(x,y;\hat{f})$ denotes the gradient aligned loss with respect to the surrogate model $\hat{f}$ achieved by the gradient aligned mechanism.

\subsection{Gradient Aligned Losses}
\label{sec:gradient_aligned_losses}
According to the gradient aligned mechanism defined in Equation~\ref{eq:gradient_aligned_mechanism}, both the CE loss and the margin loss can be transformed into the gradient aligned forms, i.e., the gradient aligned CE (GACE) loss and gradient aligned margin (GAM) loss. Specifically, the GACE loss is formulated in Equation~\ref{eq:gace} and the GAM loss in Equation~\ref{eq:gam}.
\begin{gather}
L_{GACE}(x,y;\hat{f})=\sum_i^C{\frac{\partial L_{CE}(x,y;f)}{\partial z_i^f}\cdot z_i^{\hat{f}}} \nonumber \\
=\frac{p_y^f-1}{\ln 2}\cdot z_y^{\hat{f}} + \sum_{i=1( i\ne y )}^C\frac{p_i^f}{\ln 2}\cdot z_i^{\hat{f}} \label{eq:gace} \\
p_i^f=\frac{e^{z_i^f}}{\sum\nolimits_{j=1}^C{e^{z_j^f}}}
\end{gather}
where $p_i^f$ indicates the probability of the category with index $i$ predicted by the victim model $f$.
\begin{gather}
L_{GAM}(x,y;\hat{f})=\sum_i^C{\frac{\partial L_{margin}(x,y;f)}{\partial z_i^f}\cdot z_i^{\hat{f}}} \nonumber \\
=z_i^{\hat{f}}-z_y^{\hat{f}} \label{eq:gam} \\
i=\underset{i\ne y}{arg\max}\:z_{i}^{f}
\end{gather}

We argue that the gradient aligned losses can enhance the metric of gradient alignment~\cite{gradient-alignment} between the surrogate and victim models. To demonstrate this argument, we compare the metric of gradient alignment with and without the GACE loss on several transfer attacks such as MIFGSM~\cite{MI-FGSM}, NIFGSM~\cite{SI-NI-FGSM}, VMIFGSM~\cite{VMI-FGSM}, SGM~\cite{SGM}, and MoreBayesian~\cite{MoreBayesian}. The metric of gradient alignment is quantified using Equation~\ref{eq:gradient_alignment}~\cite{DTA}.
\begin{gather}
\nu =\sum_{i=1}^{\left|S\right|}{\sum_{t=1}^T{\cos (g_{i,t}^{\hat{f}},g_{i,t}^{f})}} \label{eq:gradient_alignment}
\end{gather}
where $\left|S\right|$ is the size of the test set, $T$ is the number of iterations in the transfer attacks, $g_{i,t}^{\hat{f}}$ and $g_{i,t}^{f}$ is the gradient of $i$-th example at $t$-th iteration on the surrogate and victim models, respectively. Note that, $g_{i,t}^{\hat{f}}$ is calculated by the transfer attacks on the surrogate model and $g_{i,t}^{f}$ by a simple gradient descent method on the victim model.

\begin{figure*}[t]
\begin{center}
\centerline{\includegraphics[width=\textwidth]{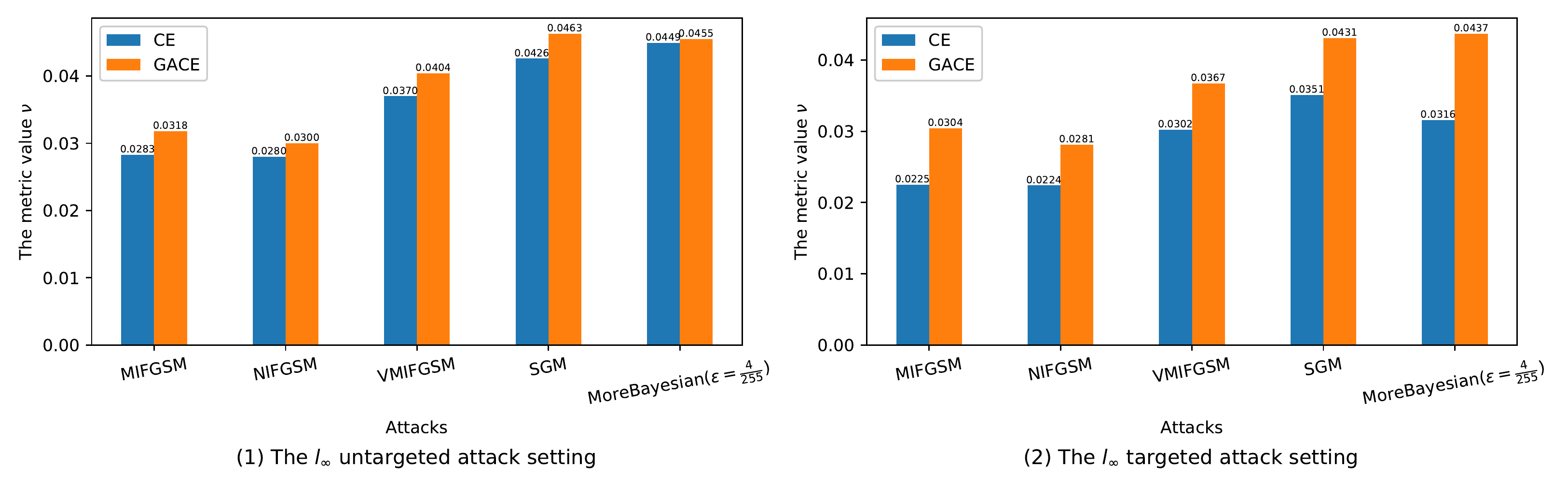}}
\caption{The metric value $\nu$ of gradient alignment on different transfer attacks with CE or GACE loss under the $l_\infty$ untargeted or targeted attack settings. The surrogate and victim models are ResNet50~\cite{ResNet} and VGG16~\cite{VGG}, respectively.}\vspace{-5mm}
\label{FIG:metric_of_gradient_alignment}
\end{center}
\end{figure*}

As shown in Fig.~\ref{FIG:metric_of_gradient_alignment}, our proposed GACE loss improves the metric value $\nu$ of gradient alignment on several recent transfer attacks under the $l_\infty$ untargeted and targeted attack settings. Furthermore, as demonstrated in Section~\ref{sec:transfer_attacks_with_GACE_loss}, the untargeted and targeted attack success rates of these transfer attacks with our GACE loss are significantly higher than those achieved using the original CE loss. These results clearly indicate that our GACE loss improves the attack performance of the transfer attacks by enhancing the metric of gradient alignment.

\begin{algorithm}[tb]
\caption{The gradient aligned attack (GAA)}
\label{alg:gaa}
\begin{algorithmic}
\Require
The surrogate model $\hat{f}$; the victim model $f$; An natural example pair $(x,y)$; the magnitude of perturbation $\epsilon$; the number of iteration $T$ and decay factor $\mu$; the maximum number of queries $\Omega$; the $l_\rho$ attack setting. Note that $T$ is greater than $\Omega$ in the proposed novel scenario.
\Ensure
An adversarial example $x^{adv}$.
\State $\alpha=\epsilon/T$
\State $g_0=0; x^{adv}_0=x$
\For {$t=0\rightarrow T-1$}
\State Query the probability vector of the victim model $f$:
\begin{align}
\label{eq:query}
\boldsymbol{p}^f=\left\{ \begin{array}{c}
f( x_{t}^{adv} ), if\ t\in \{ \lfloor \frac{T}{\Omega}\cdot i \rfloor  \mid i=0,1,\cdots, \Omega-1 \}\\
\boldsymbol{p}^f, if\ t\notin \{ \lfloor \frac{T}{\Omega}\cdot i \rfloor \mid i=0,1,\cdots, \Omega-1 \}
\end{array} \right.
\end{align}
\If{$l_\rho=l_\infty$}
\State Calculate $g_{t+1}=\mu\cdot g_t+\frac{\nabla _{x_t^{adv}}L_{GACE}( x_{t}^{adv},y;\hat{f},\boldsymbol{p}^f )}{\left\| \nabla _{x_t^{adv}}L_{GACE}( x_{t}^{adv},y;\hat{f},\boldsymbol{p}^f ) \right\|_1}$
\State Update $x^{adv}_{t+1}$ by applying the sign of gradient $x_{t+1}^{adv}=Clip_{x}^{\epsilon}( x_{t}^{adv}+\alpha \cdot sign( g_{t+1} ) )_\infty$
\ElsIf{$l_\rho=l_2$}
\State Calculate $g_{t+1}=\mu\cdot g_t+\frac{\nabla _{x_t^{adv}}L_{GACE}( x_{t}^{adv},y;\hat{f},\boldsymbol{p}^f )}{\left\| \nabla _{x_t^{adv}}L_{GACE}( x_{t}^{adv},y;\hat{f},\boldsymbol{p}^f ) \right\|_2}$
\State Update $x^{adv}_{t+1}$ by applying the gradient $x_{t+1}^{adv}=Clip_{x}^{\epsilon}( x_{t}^{adv}+\alpha \cdot g_{t+1}  )_2$
\EndIf
\EndFor
\State $x^{adv}=x^{adv}_T$\\
\Return $x^{adv}$
\end{algorithmic}
\end{algorithm}

\subsection{Gradient Aligned Attacks}
\label{sec:gradient_aligned_attacks}
In this paper, we propose the Gradient Aligned Attacks (GAA), a class of transferable prior-based query attacks that operate in a novel scenario where the number of allowed queries is limited to less than 10. These attacks are based on the proposed GACE loss and we present both $l_\infty$ and $l_2$ versions of the GAA algorithm in Algorithm~\ref{alg:gaa} when the hyperparameter $l_\rho$ is set to $l_\infty$ and $l_2$, respectively.

In Algorithm~\ref{alg:gaa}, Equation~\ref{eq:query} represents the equidistant query. The probability vector $\boldsymbol{p}^f$ also represents $[p_1^f, p_2^f,\cdots,p_C^f]$ used in Equation~\ref{eq:gace}. The $Clip_x^\epsilon(\cdot)_\rho$ function restricts the generated adversarial examples to be within the $\epsilon$-ball of $x$ under the $l_\rho$ attack setting where $\rho$ can be 2 or $\infty$.

In addition, the $l_\infty$ gradient aligned attack can be well compatible with Nesterov Accelerated Gradient (NAG)~\cite{SI-NI-FGSM}, variance-tuning method~\cite{VMI-FGSM}, skip gradient method (SGM)~\cite{SGM} and MoreBayesian method~\cite{MoreBayesian} to achieve a high attack performance, so does the $l_2$ gradient aligned attack with SGM~\cite{SGM} and MoreBayesian~\cite{MoreBayesian}. It is worth noting that when using MoreBayesian in combination with the gradient aligned attacks, the momentum term is removed.

\section{Experiments}
\label{sec:experiments}

In this section, we will introduce our experimental setup which includes the dataset used, the deep learning models, and the baselines that we have used to compare our black-box query attacks. We will also explain the hyperparameter settings that we have used for our experiments. Then the performance of our gradient aligned attacks is evaluated in comparison with the four latest transferable prior-based query attacks in the proposed novel scenario. The gradient aligned losses are demonstrated to improve the attack performance of the transfer attacks and the query efficiency of the query attacks. Finally, we will analyze the effect of the maximum queries $\Omega$ on the attack performance of the gradient aligned attacks. All experiments are evaluated on two RTX 3090s with the Pytorch toolkit.

\subsection{Setup}
\label{sec:setup}
\subsubsection{Dataset}
All of our experiments have been evaluated using a randomly selected subset of 1000 images from the ImageNet validation set~\cite{ImageNet}. We have ensured that these images are correctly classified by the deep learning models that we have used in this section.

\subsubsection{Models}

We have used six pre-trained deep learning models from the model library\footnote{https://pytorch.org/vision/stable/models.html} as our surrogate or victim models for our experiments. These models include VGG16~\cite{VGG}, VGG19~\cite{VGG}, ResNet50~\cite{ResNet}, ResNet152~\cite{ResNet}, Inception-v3~\cite{Inception-v3}, and MobileNet-v2~\cite{MobileNet}. All of these models take a $3\times 224\times 224$ dimensional image as input, and output a 1000-dimensional probability vector corresponding to 1000 different categories.

\subsubsection{Baselines}
We have selected four pure query attacks, namely Bandts-TD~\cite{Bandit}, SimBA~\cite{SimBA}, SignHunter~\cite{SignHunter}, and Square~\cite{Square}, four transferable prior-based query attacks, namely PRGF~\cite{P-RGF}, ODS~\cite{ODS}, LeBA~\cite{LeBA}, and GFCS~\cite{GFCS}, and five transfer attacks, namely MIFGSM~\cite{MI-FGSM}, NIFGSM~\cite{SI-NI-FGSM}, VMIFGSM~\cite{VMI-FGSM}, SGM~\cite{SGM}, and MoreBayesian~\cite{MoreBayesian} for our experiments. These attacks have been implemented in various code libraries\footnote{https://github.com/DIG-Beihang/AISafety/tree/main/EvalBox/Attack}\footnote{https://github.com/ermongroup/ODS}\footnote{https://github.com/TrustworthyDL/LeBA}\footnote{https://github.com/CSIPlab/BASES/tree/main/comparison/GFCS}\footnote{https://github.com/Harry24k/adversarial-attacks-pytorch/tree/master/torchattacks/attacks}.

\subsubsection{Hyperparameters}

In our proposed novel scenario, we set the maximum number of queries $\Omega$ to 5. The magnitude of the adversarial perturbation is set to $\epsilon=\frac{8}{255}$ for $l_\infty$ attacks and $\epsilon=\sqrt{0.001\times 3\times 224\times 224}\approx 12.27$ for $l_2$ attacks, aligning with the values used in previous works~\cite{P-RGF,LeBA}. To ensure adequate attack, the step size is set to $\alpha=\frac{8}{255}$ for $l_\infty$ query attacks and $\alpha=10$ for $l_2$ query attacks in this novel scenario. For our gradient aligned attacks, we set the step size to $\alpha=\frac{0.8}{255}$ under the $l_\infty$ attack setting and $\alpha=2$ under the $l_2$ attack setting, with the number of iterations set to $T=10$. For transfer attacks, we set the step size and the number of iterations to $\epsilon=\frac{0.8}{255}$ and $T=10$, respectively. However, note that for the transfer attack MoreBayesian~\cite{MoreBayesian}, we set the above parameters to $\epsilon=\frac{4}{255}$, $\alpha=\frac{1}{255}$, and $T=10$.

For the four pure query attacks, namely Bandit-TD~\cite{Bandit}, SimBA~\cite{SimBA}, SignHunter~\cite{SignHunter}, and Square~\cite{Square}, and the four transferable prior-based query attacks, namely PRGF~\cite{P-RGF}, ODS~\cite{ODS}, LeBA~\cite{LeBA}, and GFCS~\cite{GFCS}, we followed their standard setups. In Bandit-TD~\cite{Bandit}, the online learning rate, Bandit exploration, finite difference probe, and tile size are set to 100, 0.01, 0.1, and 6px under the $l_\infty$ attack setting and 0.1, 0.01, 0.01, 6px under the $l_2$ attack setting. In LeBA~\cite{LeBA}, the mode is set as "train," and the interval for TIMIFGSM attack is set to 20, where the momentum decay factor, the name of the kernel, and the length of the kernel are set to 0.9, "Gaussian," and 9 in TIMIFGSM~\cite{TI-FGSM}. In our gradient aligned attacks, the momentum decay factor is set to $\mu=1.0$.

In the five transfer attacks, we set the momentum decay factor $\mu=1.0$ for MIFGSM~\cite{MI-FGSM}, NIFGSM~\cite{SI-NI-FGSM}, VMIFGSM~\cite{VMI-FGSM}, SGM~\cite{SGM}. The number of sampled examples in the neighborhood and the upper bound of the neighborhood are set to 20 and 1.5 for VMIFGSM~\cite{VMI-FGSM}, respectively. The factor of reducing the gradient from the residual model is set to 0.2 for SGM~\cite{SGM}. The rescaling factor and the number of substitute models are set as 1.5 and 20 for MoreBayesian~\cite{MoreBayesian}.

In the transferable prior-based query attacks~\cite{P-RGF,LeBA,ODS,GFCS} and transfer attacks~\cite{MI-FGSM,SI-NI-FGSM,VMI-FGSM,SGM,MoreBayesian}, we set ResNet50 as the surrogate model when Num.S (i.e., the number of the surrogate model) is 1 and the set $\{$ResNet50, ResNet152, Inception-v3, MobileNet-v2$\}$ as the surrogate models when Num.S is 4, while VGG16 is used as the victim model. 

\subsubsection{Metrics} The untargeted/targeted attack success rate (ASR) denotes the percentage of generated adversarial examples that are misclassified as any/specified wrong category to evaluate the performance of adversarial attacks, where the target label is randomly selected under the targeted attack setting. Additionally, the average number of queries is used to evaluate the query efficiency of query attacks. Specifically, we use the Avg.Q-ASR curve to evaluate the comprehensive performance of query attacks, where Avg.Q denotes the average number of queries. Notably, the closer the Avg.Q-ASR curve is to the upper left corner, the higher the query efficiency of the query attacks, which denotes fewer queries are required on average with the same attack success rate.

\begin{table}[t]
\caption{The untargeted attack success rates of different transferable prior-based query attacks in the novel scenario. Num.S dentoes the number of surrogate models and "-"  represents the attacks is not implemented.}
\renewcommand\arraystretch{1.0}
\footnotesize
\centering
\begin{tabular}{cccc}
\hline
Attacks               & Num.S & $l_2$ attack    & $l_\infty$ attack \\ \hline
PRGF                  & 1     & 41.8\%          & 34.9\%            \\ \cline{2-2}
LeBA                  & 1     & 59.9\%          & -                 \\ \cline{2-2}
ODS                   & 1     & 62.7\%          & -                 \\
                      & 4     & 61.4\%          & -                 \\ \cline{2-2}
\multirow{2}{*}{GFCS} & 1     & 69.6\%          & 35.5\%            \\
                      & 4     & 77.4\%          & 44.1\%            \\ \cline{2-2}
\multirow{2}{*}{GAA (ours)}  & 1     & 80.7\%          & 55.9\%            \\
                      & 4     & \textbf{93.5\%} & \textbf{75.4\%}   \\ \hline
\end{tabular}
\label{tab:untargeted_comparison_of_GAA_in_the_novel_scenario}
\end{table}

\begin{table}[t]
\caption{The targeted attack success rates of different transferable prior-based query attacks in the novel scenario. Num.S dentoes the number of surrogate models.}
\renewcommand\arraystretch{1.0}
\footnotesize
\centering
\begin{tabular}{cccc}
\hline
Attacks               & Num.S & $l_2$ attack    & $l_\infty$ attack \\ \hline
\multirow{2}{*}{GFCS} & 1     & 1.1\%           & 0.3\%             \\
                      & 4     & 1.3\%           & 0.2\%             \\ \cline{2-2}
\multirow{2}{*}{GAA (ours)}  & 1     & 2.3\%           & 0.5\%             \\
                      & 4     & \textbf{13.5\%} & \textbf{3.4\%}    \\ \hline
\end{tabular}
\label{tab:targeted_comparison_of_GAA_in_the_novel_scenario}
\end{table}

\subsection{The Gradient Aligned Attacks in the Novel Scenario}
\label{sec:GAA_in_the_novel_scenario}

To evaluate the effectiveness of gradient aligned attacks, we conduct experiments to compare untargeted and targeted attack success rates with several latest transferable prior-based query attacks under $l_2$ and $l_\infty$ attack settings in our proposed novel scenario. As shown in Table~\ref{tab:untargeted_comparison_of_GAA_in_the_novel_scenario}, our gradient aligned attacks improve the untargeted attack success rate by 16.1\% under the $l_2$ attack setting and 31.3\% under the $l_\infty$ attack setting. Notably, for the ODS attack~\cite{ODS}, the untargeted attack success rate with 1 surrogate model is higher than that with 4 surrogate models in the novel scenario. This is due to the serious update oscillation caused by the large step size and randomness in each update iteration. As shown in Table~\ref{tab:targeted_comparison_of_GAA_in_the_novel_scenario}, our gradient aligned attacks improve the targeted attack success rate by 12.2\% and 3.1\% under the $l_2$ and $l_\infty$ attack settings, respectively, in comparison with the GFCS attack~\cite{GFCS}. In the novel scenario, the targeted attack success rate of the GFCS attack~\cite{GFCS} with 4 surrogate models only had a slight improvement compared to that with 1 surrogate model, while our gradient aligned attacks improve the targeted attack success rate by 11.2\% and 2.9\% under the $l_2$ and $l_\infty$ attack settings, respectively. Therefore, our gradient aligned attacks can significantly improve the attack performance in the novel scenario in comparison with the latest advanced query attacks under both the $l_2$ and $l_\infty$ attack setting, and increasing the number of the surrogate model can significantly improve the targeted attack success rate.

\begin{table}[t]
\caption{The untargeted/targeted attack success rates of different advanced transfer attacks with and without the GACE loss under the $l_\infty$ attack setting to aligned with the previous works~\cite{MI-FGSM,SI-NI-FGSM,VMI-FGSM,SGM,MoreBayesian}. Num.S denotes the number of surrogate models.}
\renewcommand\arraystretch{1.0}
\footnotesize
\centering
\resizebox{1.0\columnwidth}{!}{
\begin{tabular}{ccccc}
\hline
Attacks                                                 & Num.S              & Loss & Untargeted attack & Targeted attack \\ \hline
\multirow{4}{*}{MIFGSM}                                 & \multirow{2}{*}{1} & CE   & 47.0\%            & 0.2\%           \\
                                                        &                    & GACE & 57.1\%            & 0.5\%           \\ \cline{3-3}
                                                        & \multirow{2}{*}{4} & CE   & 71.0\%            & 1.7\%           \\
                                                        &                    & GACE & \textbf{76.7\%}   & \textbf{3.8\%}  \\ \cline{2-5} 
\multirow{4}{*}{NIFGSM}                                 & \multirow{2}{*}{1} & CE   & 54.4\%            & 0.4\%           \\
                                                        &                    & GACE & 59.3\%            & 1.0\%           \\ \cline{3-3}
                                                        & \multirow{2}{*}{4} & CE   & 75.0\%            & 2.9\%           \\
                                                        &                    & GACE & \textbf{78.2\%}   & \textbf{3.6\%}  \\ \cline{2-5} 
\multirow{4}{*}{VMIFGSM}                                & \multirow{2}{*}{1} & CE   & 65.3\%            & 1.7\%           \\
                                                        &                    & GACE & 72.5\%            & 3.3\%           \\ \cline{3-3}
                                                        & \multirow{2}{*}{4} & CE   & 78.8\%            & 4.5\%           \\
                                                        &                    & GACE & \textbf{84.6\%}   & \textbf{7.0\%}  \\ \cline{2-5} 
\multirow{2}{*}{SGM}                                    & \multirow{2}{*}{1} & CE   & 74.2\%            & 1.8\%           \\
                                                        &                    & GACE & \textbf{78.2\%}   & \textbf{3.9\%}  \\ \cline{2-5} 
\multirow{2}{*}{MoreBayesian($\epsilon=\frac{4}{255}$)} & \multirow{2}{*}{1} & CE   & 76.1\%            & 6.9\%           \\
                                                        &                    & GACE & \textbf{84.2\%}   & \textbf{12.0\%} \\ \hline
\end{tabular}
}
\label{tab:transfer_attacks_with_GACE_loss}
\end{table}

\subsection{The Transfer Attacks with the Gradient Aligned CE Loss}
\label{sec:transfer_attacks_with_GACE_loss}

In this section, we evaluate the effectiveness of the GACE loss in improving the attack performance of transfer attacks. We conduct untargeted and targeted attack experiments on five advanced transfer attacks~\cite{MI-FGSM,SI-NI-FGSM,VMI-FGSM,SGM,MoreBayesian} with and without the GACE loss under the $l_\infty$ attack setting, to align with the latest works. Due to the surrogate model needing to be a ResNet-like model, we only consider the 1 surrogate model setting in the SGM~\cite{SGM} attack. Moreover, as the surrogate models have high attack performance, we reduce the magnitude of the adversarial perturbations to $\frac{4}{255}$. Our results in Table~\ref{tab:transfer_attacks_with_GACE_loss} show that the GACE loss can improve the untargeted attack success rate by 3.2 to 10.1\% and the targeted attack success rate by 0.3 to 5.1\% on the five latest advanced transfer attacks. Additionally, as shown in Fig.~\ref{FIG:metric_of_gradient_alignment}, our evaluation of the metric value of gradient alignment on these attacks with the GACE loss, is higher than that with the CE loss. Thus, our GACE loss can enhance the attack performance of transfer attacks by increasing the metric value of gradient alignment in the untargeted and targeted attacks.

\begin{figure}[t]
\begin{center}
\centerline{\includegraphics[width=\columnwidth]{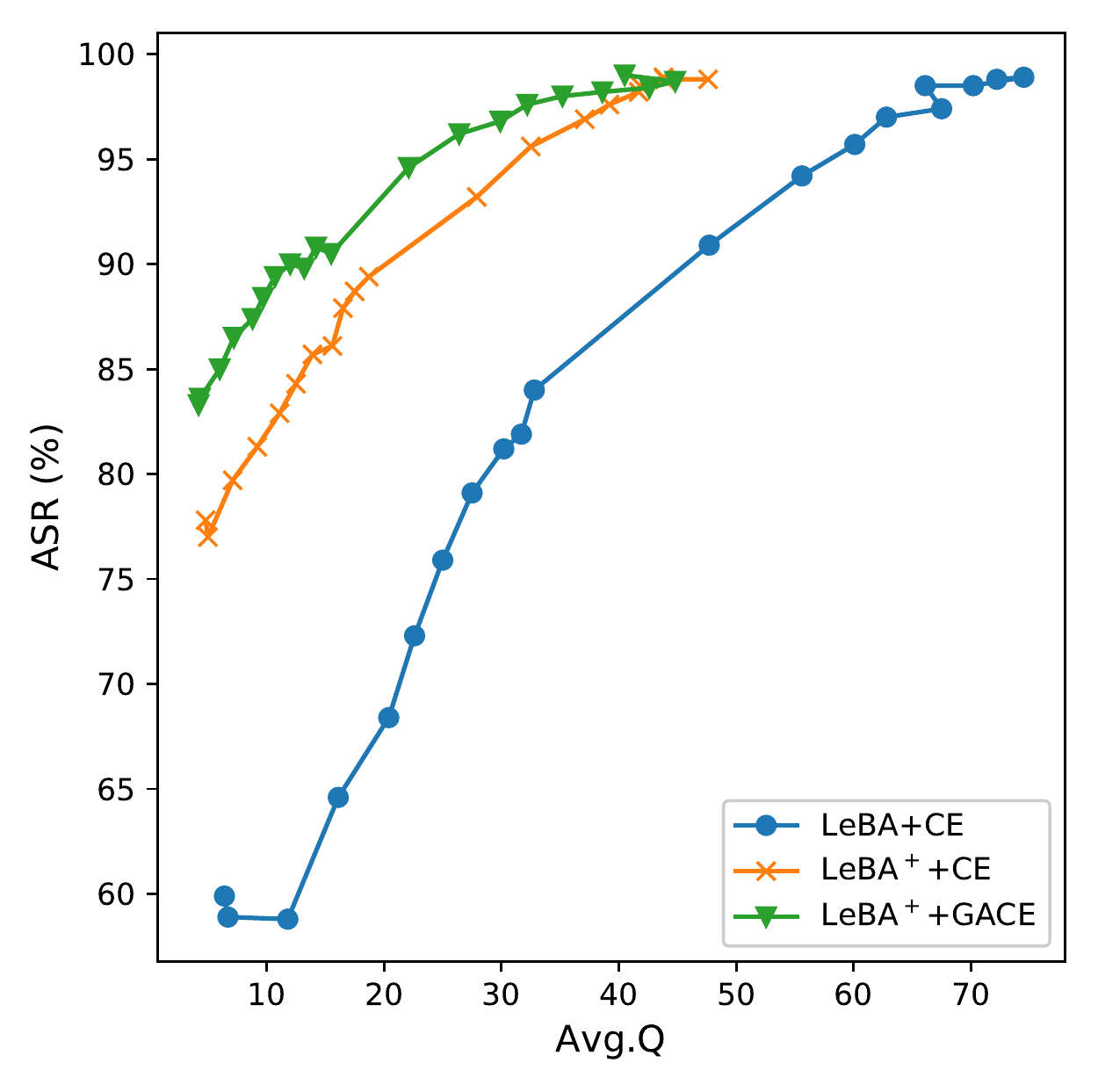}}
\caption{The Avg.Q-ASR (\%) curves of LeBA~\cite{LeBA} and its improved version by us, namely LeBA$^+$, with or without the GACE loss under the $l_2$ untargeted attack setting.}\vspace{-5mm}
\label{FIG:LeBA_with_GAL}
\end{center}
\end{figure}

\begin{figure}[t]
\begin{center}
\centerline{\includegraphics[width=\columnwidth]{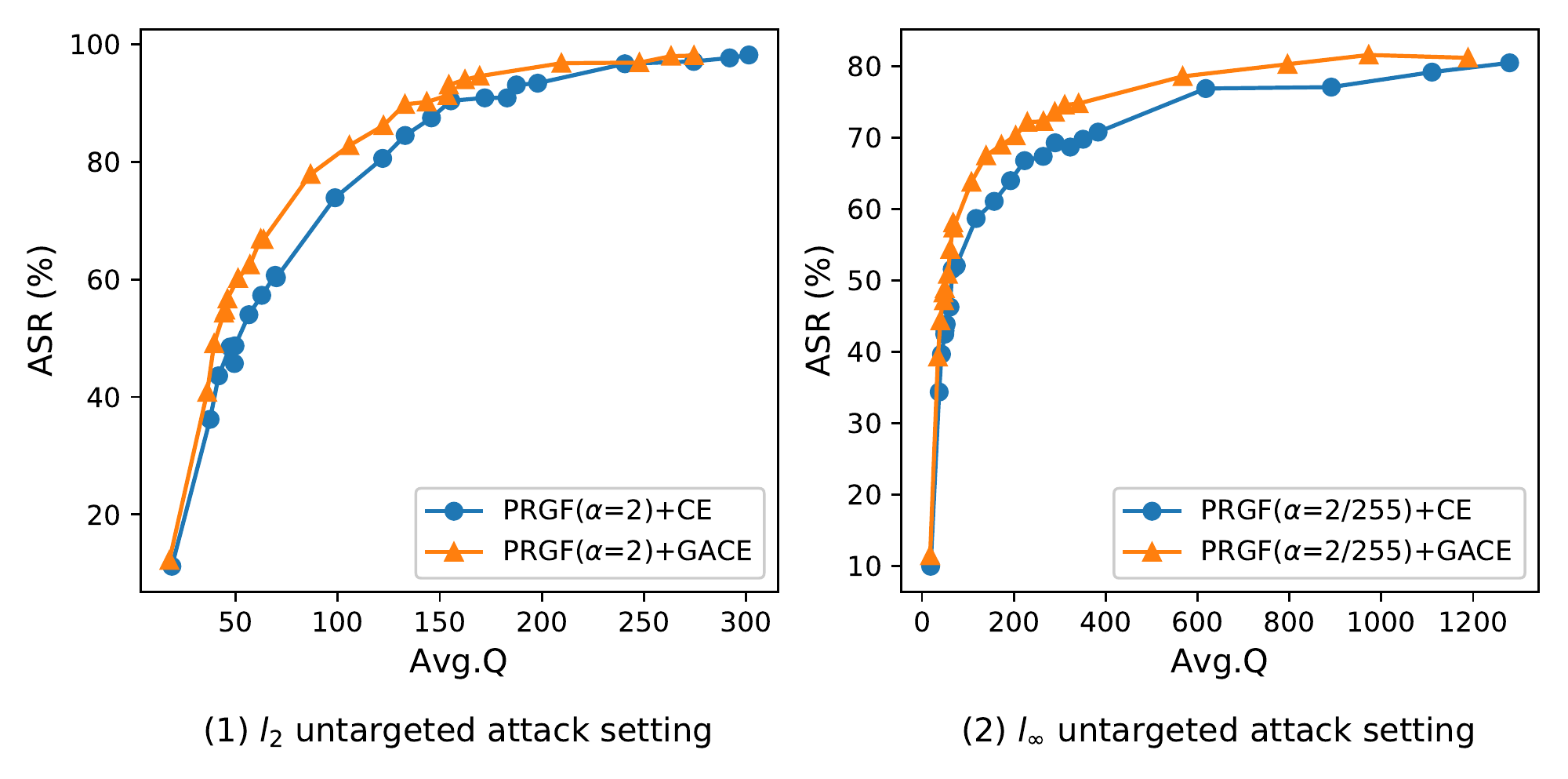}}
\caption{The Avg.Q-ASR (\%) curves of PRGF~\cite{P-RGF} with or without the GACE loss under the $l_2$ and $l_\infty$ untargeted attack settings.}\vspace{-5mm}
\label{FIG:PRGF_with_GAL}
\end{center}
\end{figure}

\begin{figure}[t]
\begin{center}
\centerline{\includegraphics[width=\columnwidth]{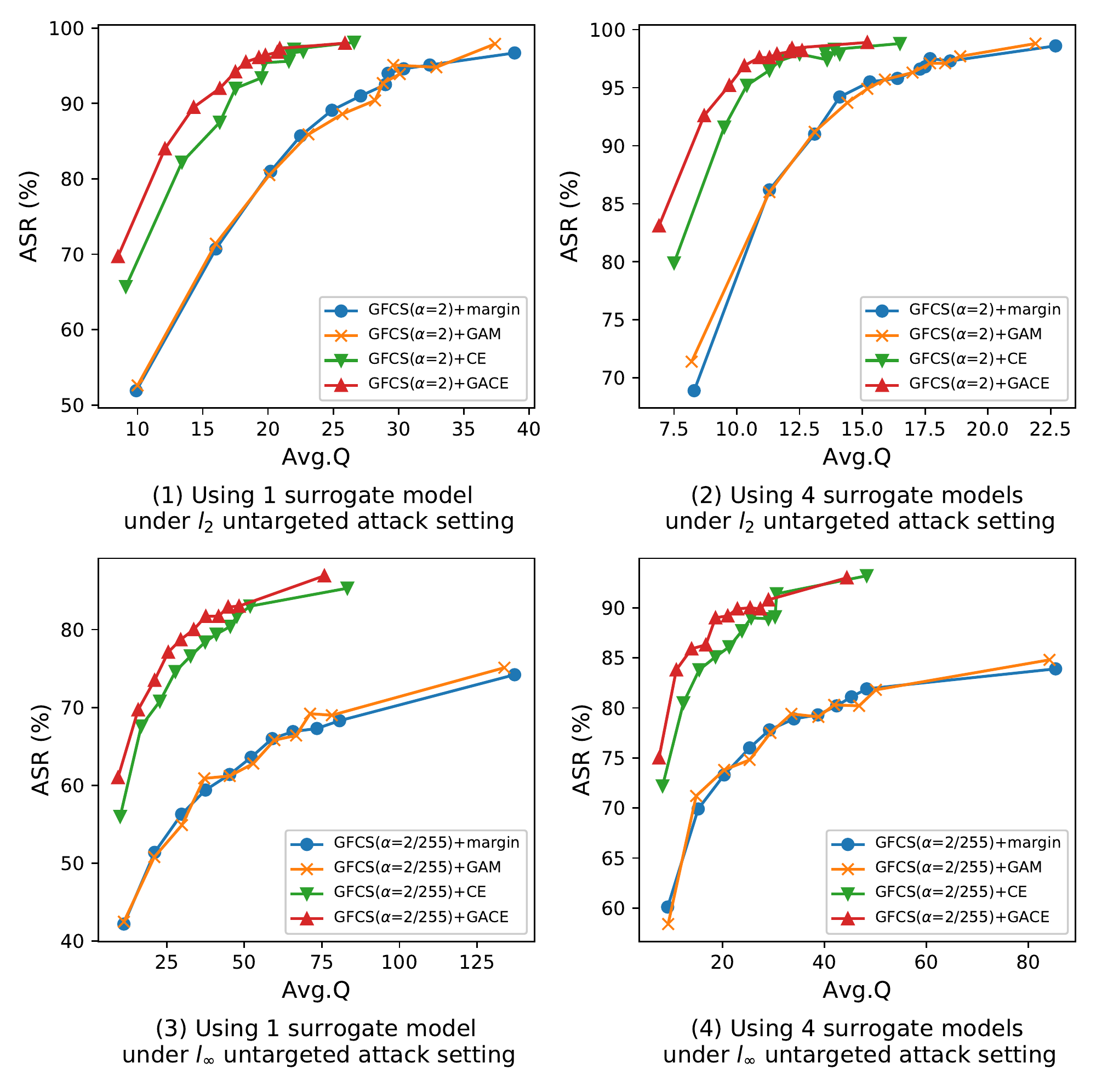}}
\caption{The Avg.Q-ASR (\%) curves of GFCS~\cite{GFCS} with or without the GAM or GACE loss using 1 and 4 surrogate models under the $l_2$ and $l_\infty$ untargeted attack settings.}\vspace{-5mm}
\label{FIG:GFCS_with_GAL_untargeted}
\end{center}
\end{figure}

\begin{figure}[t]
\begin{center}
\centerline{\includegraphics[width=\columnwidth]{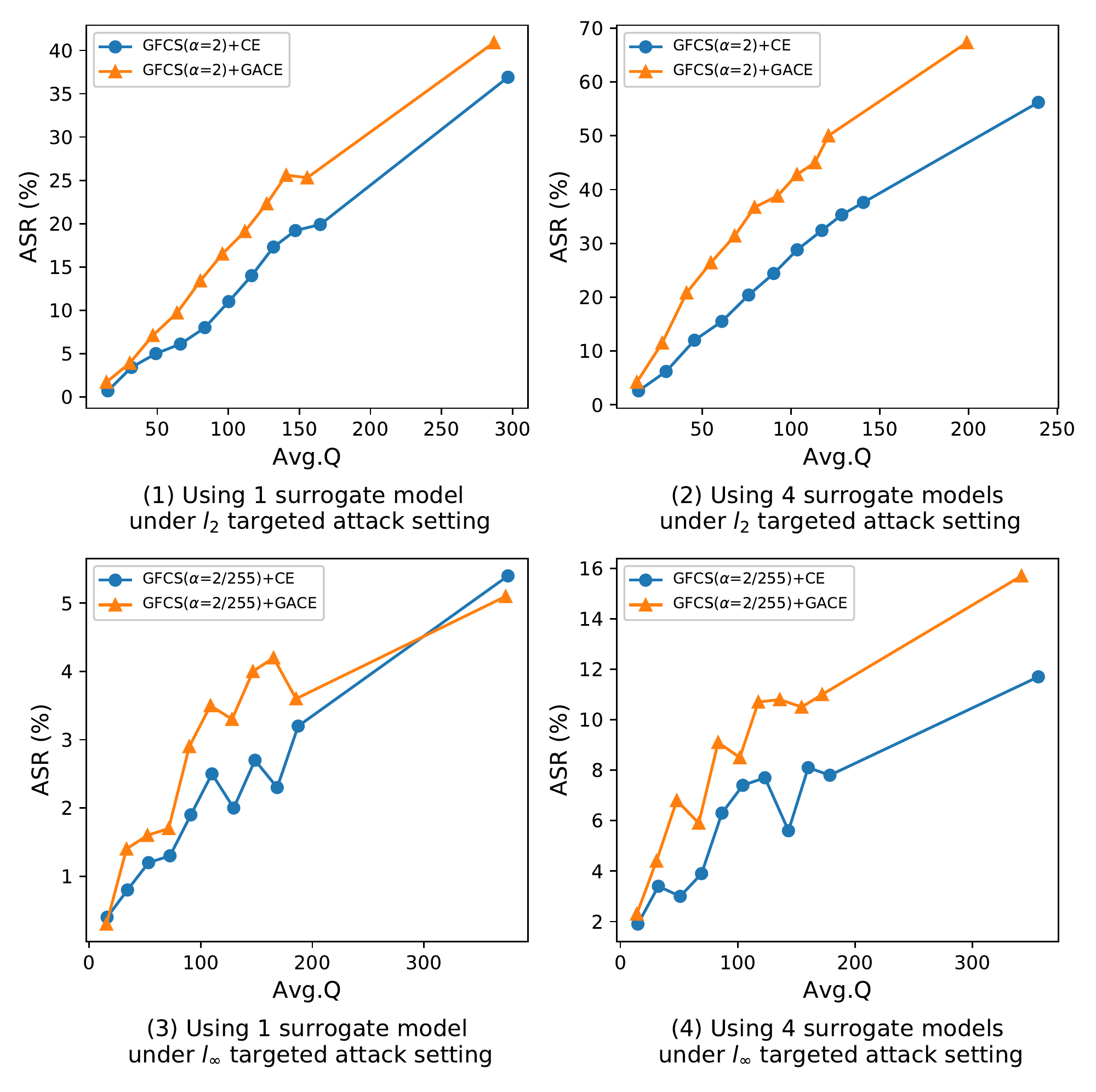}}
\caption{The Avg.Q-ASR (\%) curves of GFCS~\cite{GFCS} with or without the GACE loss using 1 and 4 surrogate models under the $l_2$ and $l_\infty$ targeted attack settings.}\vspace{-5mm}
\label{FIG:GFCS_with_GAL_targeted}
\end{center}
\end{figure}

\subsection{The Query Attacks with the Gradient Aligned Losses}
\label{sec:query_attacks_with_GAL}

In this section, we evaluate the effectiveness of our gradient-aligned losses in improving the query efficiency of the latest advanced transferable prior-based query attacks such as LeBA~\cite{LeBA}, PRGF~\cite{P-RGF}, and GFCS~\cite{GFCS}, using their standard setups. The step size for PRGF~\cite{P-RGF} and GFCS~\cite{GFCS} is set to $2$ and $\frac{2}{255}$, respectively, for the $l_2$ and $l_\infty$ attack settings. Additionally, we improve the LeBA~\cite{LeBA} attack by replacing the update gradient formula of Equation~\ref{eq:update_gradient_formula_of_LeBA} with that of Equation~\ref{eq:update_gradient_formula_of_LeBA_plus} in the TIMIFGSM part of LeBA. We refer to this improved version as LeBA$^+$. Notably, the original LeBA and PRGF use the CE loss, and the original GFCS uses the margin loss in the untargeted attack and the CE loss in the targeted attack. To evaluate the effectiveness of the gradient-aligned losses, we compare the GACE loss with the CE loss and the GAM loss with the margin loss, respectively.
\begin{gather}
g_{t+1}=\frac{\mu\cdot g_t+\nabla _{x_t^{adv}}L( x_{t}^{adv},y)}{\left\| \mu\cdot g_t + \nabla _{x_t^{adv}}L( x_{t}^{adv},y) \right\|_2} \label{eq:update_gradient_formula_of_LeBA} \\
g_{t+1}=\mu\cdot g_t+\frac{\nabla _{x_t^{adv}}L( x_{t}^{adv},y)}{\left\| \nabla _{x_t^{adv}}L( x_{t}^{adv},y) \right\|_2} \label{eq:update_gradient_formula_of_LeBA_plus}
\end{gather}
%


As shown in Fig.~\ref{FIG:LeBA_with_GAL}, we conduct experiments to compare the Avg.Q-ASR (\%) curves of LeBA~\cite{LeBA} and LeBA$^+$ with and without the GACE loss under the $l_2$ untargeted attack setting. The curves are plotted by varying the maximum number of queries from 10 to 1000. The results in Fig.~\ref{FIG:LeBA_with_GAL} show that the curve of LeBA$^+$ is closer to the upper-left corner than the curve of LeBA, which demonstrates the effectiveness of LeBA$^+$ in improving the query efficiency. Furthermore, the GACE loss can improve the query efficiency of LeBA$^+$ by a maximum factor of 2.9 times under the same attack performance. Therefore, LeBA$^+$ is a successful improvement of LeBA, and the GACE loss can further improve the query efficiency of LeBA$^+$.

As shown in Fig.~\ref{FIG:PRGF_with_GAL}, we compare the Avg.Q-ASR (\%) curves between PRGF~\cite{P-RGF} with and without the GACE loss under the $l_2$ and $l_\infty$ untargeted attack settings. The curves are plotted by varying the maximum number of queries from 10 to 5000. The results show that PRGF with the GACE loss has a higher query efficiency than without the GACE loss, as it is closer to the upper left corner in the Avg.Q-ASR (\%) coordinates. Quantitatively, the query efficiency improves by a maximum factor of 1.4 and 1.9 times under the $l_2$ and $l_\infty$ attack settings, respectively. Therefore, our GACE loss can improve the query efficiency of PRGF~\cite{P-RGF} by replacing the original CE loss.

As shown in Fig.~\ref{FIG:GFCS_with_GAL_untargeted} and Fig.~\ref{FIG:GFCS_with_GAL_targeted}, we present a systematic evaluation of gradient aligned losses, including the GACE and GAM losses, on positively affecting the query efficiency of GFCS~\cite{GFCS} under different settings. These settings include various combinations of the number of surrogate models, attack types (i.e., $l_2$ or $l_\infty$), and targeted or untargeted attacks. We also explore the effectiveness of the GACE and CE losses on the query efficiency of GFCS under the untargeted attack setting. The curves are plotted by varying the maximum number of queries from 20 to 400. Our results demonstrate that the GACE loss leads to the best query efficiency on GFCS and improves this metric by a maximum factor of 1.2 and 1.4 times in comparison with the CE loss under the $l_2$ and $l_\infty$ untargeted attack settings, respectively. Moreover, under the targeted attack setting, the GACE loss can significantly improve the query efficiency of GFCS by a maximum factor of 1.8 and 2.2 times in comparison with the CE loss under the $l_2$ and $l_\infty$ settings, respectively. Furthermore, our study shows that the GFCS with the GAM loss has higher query efficiency than the attack with the margin loss when the maximum number of queries is large. Overall, our gradient aligned losses show a positive effect on improving the query efficiency of GFCS under different settings, including various combinations of the number of surrogate models, attack types, and targeted or untargeted attacks.

Overall, our experiments demonstrate that the gradient aligned losses, such as the GACE and GAM losses, can significantly improve the query efficiency of state-of-the-art transferable prior-based query attacks. Specifically, our approach improves the query efficiency by a maximum factor of 1.2 to 2.9 times under both $l_2$ and $l_\infty$ untargeted/targeted attack settings. These results suggest that our gradient aligned losses can be a promising approach to improve the query efficiency of various transfer-based attack methods in practical applications.

\begin{figure}[t]
\begin{center}
\centerline{\includegraphics[width=\columnwidth]{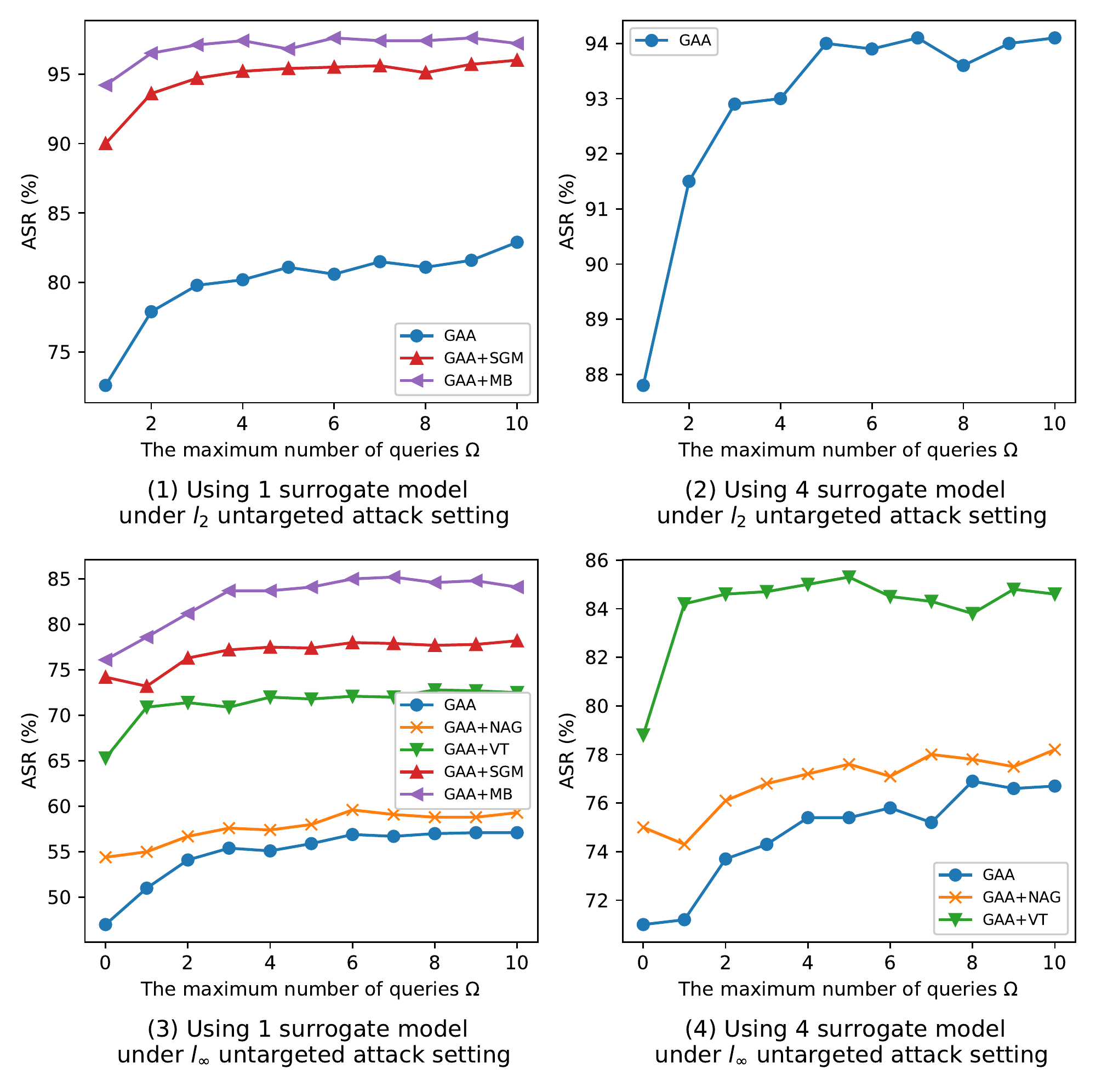}}
\caption{The untargeted attack success rate of the gradient aligned attack and its variants using 1 and 4 surrogate models under the $l_2$ and $l_\infty$ untargeted attack settings when varying the maximum number of queries $\Omega$.}\vspace{-5mm}
\label{FIG:ablation_omega_untargeted}
\end{center}
\end{figure}

\begin{figure}[t]
\begin{center}
\centerline{\includegraphics[width=\columnwidth]{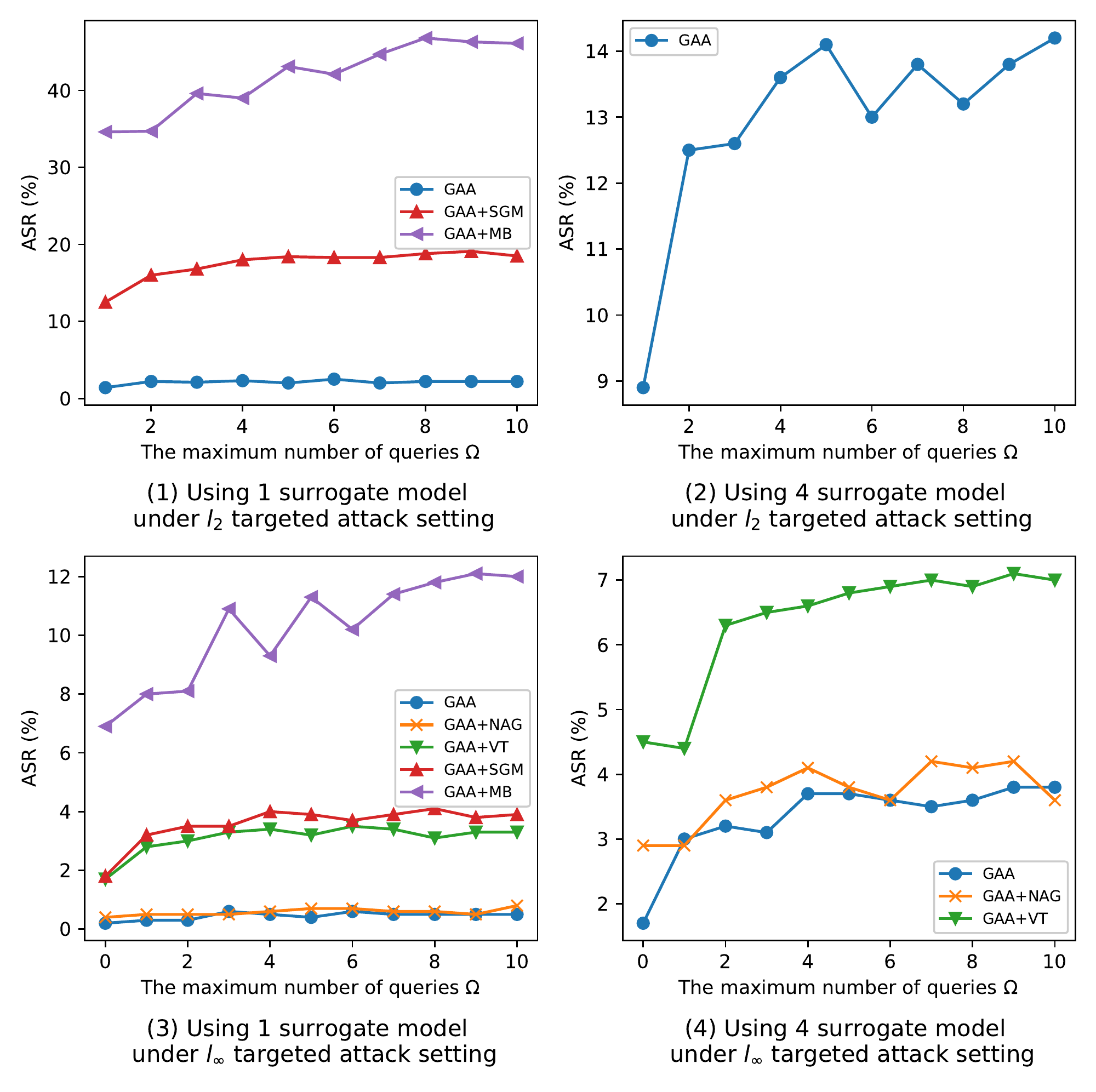}}
\caption{The targeted attack success rate of the gradient aligned attack and its variants using 1 and 4 surrogate models under the $l_2$ and $l_\infty$ targeted attack settings when varying the maximum number of queries $\Omega$.}\vspace{-5mm}
\label{FIG:ablation_omega_targeted}
\end{center}
\end{figure}

\subsection{The Sensitivity Analysis of Hyperparameters}
\label{sec:sensitivity_analysis_of_hyperparameters}

In this section, we analyze the sensitivity of the gradient aligned attack and its variants to the maximum number of queries $\Omega$, in terms of the attack success rate, under different attack settings. These settings include variations in the number of surrogate models (i.e., 1 or 4), attack types (i.e., $l_2$ or $l_\infty$), and targeted or untargeted attacks. We also explore the performance of the gradient aligned attack variants, which incorporate methods such as NAG~\cite{SI-NI-FGSM}, variance-tuning (VT)~\cite{VMI-FGSM}, SGM~\cite{SGM}, and MoreBayesian (MB)~\cite{MoreBayesian}, to stabilize the update direction or improve the accuracy of the estimated gradient. It's important to note that, under the $l_\infty$ attack setting, setting the maximum number of queries to $\Omega=0$ leads to the gradient aligned attack and its variants being degenerated to MIFGSM~\cite{MI-FGSM} and its variants, by replacing the GACE loss with the CE loss in Algorithm~\ref{alg:gaa}.


Fig.~\ref{FIG:ablation_omega_untargeted} and Fig.~\ref{FIG:ablation_omega_targeted} respectively show the attack success rates of the gradient aligned attack and its variants under the untargeted and targeted attack settings. Specifically, the results show that: i) the attack success rates of the gradient aligned attack and its variants increase gradually as $\Omega$ increases, especially fast in the range $\Omega\in[1,5]$. ii) The variants of the gradient aligned attack significantly increase the attack success rate of the gradient aligned attack. iii) As $\Omega$ increases, the targeted attack success rates of the variants of the gradient aligned attack increase more than that of the gradient aligned attack shown in Fig.~\ref{FIG:ablation_omega_targeted}. iv) As shown in Fig.~\ref{FIG:ablation_omega_untargeted}-(3),(4) and Fig.~\ref{FIG:ablation_omega_targeted}-(3),(4), the attack success rates of the gradient aligned attack and its variants are greater than that of MIFGSM~\cite{MI-FGSM} and its variants under the $l_\infty$ attack setting where the setup is consistent with the previous works~\cite{MI-FGSM,SI-NI-FGSM,VMI-FGSM}. Therefore, these analyses demonstrate that our gradient aligned attack and its variants exhibit the best attack performance in comparison with other query attacks~\cite{P-RGF,LeBA,ODS,GFCS} in the proposed novel scenario with $\Omega=5$. Moreover, as $\Omega$ increases, our proposed gradient aligned mechanism plays a significant positive role in improving the attack performance of the gradient aligned attack and its variants.

\begin{table*}[t]
\caption{The untargeted attack success rates of the advanced transfer attacks and our variants of the gradient aligned attack under the $l_\infty$ attack setting. Note that the adversarial examples are generated by the surrogate model, ResNet50~\cite{ResNet} in the transfer attacks and our query attacks.}
\renewcommand\arraystretch{1.0}
\footnotesize
\centering
\begin{tabular}{cccccc}
\hline
\multirow{2}{*}{Attacks} & \multicolumn{4}{c}{Other victim models}                             & \multirow{2}{*}{\textbf{Average}} \\ \cline{2-5}
                         & VGG19         & ResNet152     & Inception-v3  & MobileNet-v2  &                                   \\ \hline
VMIFGSM                  & 63.0          & 79.8          & 50.2          & 66.9          & 65.0                              \\
GAA+VT                   & \textbf{66.2} & \textbf{81.5} & \textbf{50.3} & \textbf{70.2} & \textbf{67.1}                     \\ \hline
MoreBayesian ($\epsilon=\frac{4}{255}$)             & 71.1          & 80.2          & 40.4          & 79.0          & 67.7                              \\
GAA+MB ($\epsilon=\frac{4}{255}$)                   & \textbf{78.8} & \textbf{82.1} & \textbf{46.6} & \textbf{81.6} & \textbf{72.3}                     \\ \hline
\end{tabular}
\label{tab:transferability_comparison}
\end{table*}

\subsection{The Transferability of the Gradient Aligned Attacks}
\label{sec:transferability_of_GAA}

In this section, we demonstrate that certain variants of the gradient aligned attack, such as "GAA+VT" and "GAA+MB", are capable of generating adversarial examples with high transferability. Transferability refers to the ability of the adversarial examples generated by the attack to successfully transfer to other victim models. Notably, when $\Omega=0$, the variants "GAA+VT" and "GAA+MB" are equivalent to VMIFGSM~\cite{VMI-FGSM} and MoreBayesian~\cite{MoreBayesian}, respectively, as the GACE loss is replaced with the CE loss in Algorithm~\ref{alg:gaa}.


As shown in Table~\ref{tab:transferability_comparison}, we compare the transferability of the adversarial examples generated by the advanced transfer attacks and our variants of the gradient aligned attack on four other victim models. The results demonstrate that the average attack success rate can be improved by 2.1-4.6\%. Therefore, our variants of the gradient aligned attacks, such as "GAA+VT" and "GAA+MB", can generate adversarial examples with higher transferability than advanced transfer attacks such as VMIFGSM~\cite{VMI-FGSM} and MoreBayesian~\cite{MoreBayesian}.

\section{Related Work}
\label{sec:related_work}

\subsection{Adversarial Attacks}
\label{sec:adversarial_attacks}

\subsubsection{White-box Attacks}
The white-box attacks know the architecture and parameters of the deep learning model. This type of attack can directly optimize the adversarial examples through the victim model according to the specific target. For gradient iterative-based attacks such as PGD~\cite{Madry-AT} and BIM~\cite{I-FGSM}, the target of them is to maximize the CE loss by adding the perturbation into the input for determining the low probability of the correct label. The optimization-based attacks such as CW~\cite{CW} and DeepFool~\cite{DeepFool} are targeted by decreasing the confidence of the ground truth and minimizing the added perturbation. The sparse attacks such as JSMA~\cite{JSMA} and Pixel attack~\cite{Pixel-attack} are $l_0$ attacks, which perturbed the important pixel to reduce the confidence of the ground truth. Recently, a strong combination attack method, namely AutoAttack~\cite{AutoAttack}, is proposed combining the APGD-CE, APGD-T, FAB-T and Square~\cite{Square} attacks. This attack has become standard to evaluate the effectiveness of the adversarial training methods~\cite{Madry-AT,Free-AT,Fast-AT,TRADES}.

\subsubsection{Black-box Attacks}
This type of attack can be divided into two categories: transfer attacks and query attacks. Both of them differ in the amount of information obtained from the victim model. Specifically, the transfer attacks do not require any information about the victim model itself, including its output, but instead need access to the training database or its related database. The query attacks require querying the output of the victim model to estimate the gradient and determine whether the attack succeeds.

\textbf{Transfer Attacks.} Currently, the transfer attacks are classified as five types, including the gradient iterative-based attacks~\cite{I-FGSM,MI-FGSM,SI-NI-FGSM,VMI-FGSM,RCE,RAP,DTA}, intermediate feature perturbation-based attacks~\cite{FIA,NAA,ATA,AA,FDA+xent,ILA,FDA,ILA++}, gradient smooth methods~\cite{LinBP,SGM}, model augmentation methods~\cite{LGV,MoreBayesian} and data augmentation~\cite{DI-FGSM,SI-NI-FGSM,TI-FGSM,Admix}. 

\textit{The gradient iterative-based attacks} improved the optimization algorithms of generating adversarial examples by stabilizing the update direction for avoiding the bad local optimum. Recently, Qin et al.~\cite{RAP} changed the max optimization as the min-max optimization forms to find the flat maximum, which has robust transferability to the victim model. 

\textit{The intermediated feature perturbation-based attacks}~\cite{AA,FIA} believed the intermediate features are more likely to be model-independent than the last layer of features due to the overfitting that happened in the last layer, thereby improving the transferability of the generated adversarial examples. In addition, this type of transfer attack focused on designing methods to accurately evaluate the importance of each intermediate feature~\cite{AA,FIA,NAA}. 

\textit{The gradient smooth methods} concluded the drawbacks of the architecture of the surrogate model such as the relu activation and residual block, resulting in the inaccurate gradient of the input. Therefore, LinBP~\cite{LinBP} removed the relu activations of the surrogate model in the gradient backpropagation and SGM reduced the gradient backpropagated from the residual block. 

\textit{The model augmentation methods} such as LGV~\cite{LGV} and MoreBayesian~\cite{MoreBayesian} are the best effective methods recently by enhancing the diversity of the surrogate model using fine-tuning method. 

\textit{The data augmentation methods} used the invariant properties of the input transformation such as padding and resize~\cite{DI-FGSM}, scale invariant~\cite{SI-NI-FGSM} and translation invariant~\cite{TI-FGSM} to perturb the common features, which are attention by all kinds of deep learning models. 

However, these attacks do not consider to combining with the output of the victim model when a few queries are allowed.

\textbf{Query Attacks.} According to the achieved amount of information from the victim model, the query attacks are further divided into hard label-based query attacks~\cite{boundary-attack,opt-attack,Sign-opt,HopSkipJumpAttack,Latent-HSJA} and score-based query attacks~\cite{ZOO,NES,Bandit,SimBA,SignHunter,Square,P-RGF,LeBA,TREMBA,ODS,GFCS} where the former only knows the predicted label of the victim model to the query but the latter knows the probability vector. 

On one hand, in the hard label-based query attacks, boundary attack~\cite{boundary-attack} and hop skip jump attack~\cite{HopSkipJumpAttack} minimized the perturbation using the boundary search method from an initial adversarial example with a large perturbation, while OPT attack~\cite{opt-attack} and Sign-OPT~\cite{Sign-opt} reformulated the boundary search method as the continuous real-value optimization targeted with minimizing the magnitude of the perturbation from the original example. Moreover, Latent-HSJA~\cite{Latent-HSJA} is a decision-based unrestricted black-box attack on facial recognition task where the unrestricted attack changes the semantic information of the input such as color schemes~\cite{NCF} and rotation of objects without perceived by humans. 

On another hand, in the score-based query attacks, the pure query attacks~\cite{ZOO,NES,Bandit,SimBA,SignHunter,Square} were proposed firstly by simplifying the optimization algorithm gradually to reduce the number of queries. Besides, Ilyas et al.~\cite{Bandit} tried to find the prior knowledge such as time-dependent and data-dependent priors to simplify the gradient estimation by reducing the search space. Then, inspired by using the priors to increase the query efficiency, the transferable priors from the surrogate model are utilized to compress the search space for increasing the query efficiency, which is called transferable prior-based query attacks~\cite{P-RGF,TREMBA,LeBA,ODS,GFCS}. Specifically, PRGF~\cite{P-RGF} combined the surrogate priors and random gradient-free method by a theoretically verified weight coefficient to increase the accuracy of the gradient estimation. LeBA~\cite{LeBA} proposed high-order gradient approximation to update the surrogate model using the query feedbacks, thereby improving the accuracy of the gradient estimation to the unsuccessfully attacked examples by the transfer attacks~\cite{TI-FGSM}. TREMBA~\cite{TREMBA} encoded the input into a low-dimensional latent vector to reduce the search space exponentially and also used the surrogate prior so that the combination of them can improve the query efficiency. ODS~\cite{ODS} improves the transferability of the estimated gradient by the surrogate model by maximizing its output diversity, but leads to a possible problem of the update oscillation. To solve the update oscillation, GFCS~\cite{GFCS} reduced the frequency of using the output diversification method by replacing it with the descent gradient of the surrogate model. The method of GFCS~\cite{GFCS} is effective to reduce the median queries to less than 10.

However, in the proposed novel scenario where the maximum number of queries is less than 10, the above advanced query attacks exhibit a low attack performance.

In addition to the image classification task, adversarial attacks are not only used to destroy the objection detection~\cite{GARSDC}, speech recognition~\cite{Feature-level-transformation} and natural language processing systems~\cite{Text-Fooler,Bert-attack}, phishing detectors~\cite{POC} and learning-based grid voltage stability assessment~\cite{GVSA}, but also protect the user privacy~\cite{UniAP}.

\subsection{Adversarial Defenses}
\label{sec:adversarial_defenses}

\subsubsection{Adversarial Training}
The adversarial training (AT) algorithms are the best effective methods to defend against the adversarial examples by using the adversarial examples as the training data of the deep learning model. Specifically, Madry et al.~\cite{Madry-AT} proposed the vanilla AT method, namely Madry-AT, which used PGD attacks to generate adversarial examples in each iteration. Then, Free-AT~\cite{Free-AT}, YOPO~\cite{YOPO}, Fast-AT~\cite{Fast-AT} and ATTA~\cite{ATTA} are proposed to accelerate the process of adversarial training. Due to the large decline of the generalization in adversarial training, TRADES~\cite{TRADES} proposed to trade off the generalization and robustness of the deep learning model and mitigated the problem of adversarial training overfit. FAT~\cite{FAT} and LAS-AT~\cite{LAS-AT} are focused on generating effective adversarial examples to mitigate the conflicts between the adversarial examples and natural examples. Besides, the data augmentation methods~\cite{Fixing-data,AdvAug,AugMix,RandAugment,NoisyMix,Proxy-distributions} are the best effective method to address the adversarial training overfit using more training data from extra dataset~\cite{Fixing-data} or generated by the input transformations~\cite{AugMix,RandAugment,NoisyMix} and generated models~\cite{Proxy-distributions}. In addition, several studies~\cite{CARDs,RobustWRN,RobustResNets} explored the robust generalization of adversarial training from the perspective of network architecture including the width, depth and capacity of the deep learning model.

\subsubsection{Adversarial Detection}
Adversarial detection algorithms are a type of method to reject adversarial examples, thereby reducing the negative effects to deep learning-based applications. Currently, the advanced detection algorithms are mainly divided into input processing methods~\cite{Feature-Squeezing,MPD,ADDITION} and statistical methods~\cite{LID,MD,KDBU,NIC,Adversarial-cone,LiBRe,JTLA} for the intermediate features of deep learning models. Specifically, the input processing methods~\cite{Feature-Squeezing,MPD,ADDITION} reconstructed the input to restore the correct prediction due to the size and direction sensitivities of the added perturbation. The statistical methods~\cite{LID,MD,KDBU,NIC,Adversarial-cone,LiBRe,JTLA} are based on the differences between the natural examples and adversarial examples in high-level representations of the victim model to detect the abnormal one. 

Recently, Sheikholeslami et al.~\cite{Provably-robust-classification} proposed a novel defense method that involves jointly training a robust classifier and detector. This approach aims to correctly classify or detect adversarial examples. Furthermore, Tram{\`{e}}r~\cite{Hardness-Reduction} argued that adversarial detection is as hard as adversarial classification (i.e., correctly classifying adversarial examples) and presented a hardness reduction showing that an adversarial detector for $\epsilon$ magnitude of adversarial perturbations can achieve the same performance as an adversarial classifier against $\frac{\epsilon}{2}$ magnitude of adversarial perturbations.

\section{Conclusion}
\label{sec:conclusion}
In this paper, we observed that current query attacks show low attack success rates in a novel scenario where only a few queries are allowed. Therefore, we propose gradient aligned attacks to enhance the best attack performance under the $l_2$ and $l_\infty$ untargeted/targeted attack settings in this scenario. Furthermore, the proposed gradient aligned losses not only improve the attack performance of the latest transfer attacks but also effectively enhance the query efficiency of advanced transferable prior-based query attacks under untargeted and targeted attack settings. In addition, the adversarial examples generated by the variants of gradient aligned attacks such as "GAA+VT" and "GAA+MB" have good transferability. Overall, our proposed gradient aligned attacks and losses show significant improvements in the attack performance and query efficiency of black-box query attacks, particularly in scenarios where only a few queries are allowed.


\bibliographystyle{IEEEtran}
\bibliography{ref}

\begin{thebibliography}{10}
\providecommand{\url}[1]{#1}
\csname url@samestyle\endcsname
\providecommand{\newblock}{\relax}
\providecommand{\bibinfo}[2]{#2}
\providecommand{\BIBentrySTDinterwordspacing}{\spaceskip=0pt\relax}
\providecommand{\BIBentryALTinterwordstretchfactor}{4}
\providecommand{\BIBentryALTinterwordspacing}{\spaceskip=\fontdimen2\font plus
\BIBentryALTinterwordstretchfactor\fontdimen3\font minus
  \fontdimen4\font\relax}
\providecommand{\BIBforeignlanguage}[2]{{%
\expandafter\ifx\csname l@#1\endcsname\relax
\typeout{** WARNING: IEEEtran.bst: No hyphenation pattern has been}%
\typeout{** loaded for the language `#1'. Using the pattern for}%
\typeout{** the default language instead.}%
\else
\language=\csname l@#1\endcsname
\fi
#2}}
\providecommand{\BIBdecl}{\relax}
\BIBdecl

\bibitem{VGG}
K.~Simonyan and A.~Zisserman, ``Very deep convolutional networks for
  large-scale image recognition,'' in \emph{3rd International Conference on
  Learning Representations, {ICLR} 2015, San Diego, CA, USA, May 7-9, 2015,
  Conference Track Proceedings}, 2015.

\bibitem{ResNet}
K.~He, X.~Zhang, S.~Ren, and J.~Sun, ``Deep residual learning for image
  recognition,'' in \emph{2016 {IEEE} Conference on Computer Vision and Pattern
  Recognition, {CVPR} 2016, Las Vegas, NV, USA, June 27-30, 2016}.\hskip 1em
  plus 0.5em minus 0.4em\relax {IEEE} Computer Society, 2016, pp. 770--778.

\bibitem{Inception-v3}
C.~Szegedy, V.~Vanhoucke, S.~Ioffe, J.~Shlens, and Z.~Wojna, ``Rethinking the
  inception architecture for computer vision,'' in \emph{2016 {IEEE} Conference
  on Computer Vision and Pattern Recognition, {CVPR} 2016, Las Vegas, NV, USA,
  June 27-30, 2016}.\hskip 1em plus 0.5em minus 0.4em\relax {IEEE} Computer
  Society, 2016, pp. 2818--2826.

\bibitem{DenseNet}
G.~Huang, Z.~Liu, L.~van~der Maaten, and K.~Q. Weinberger, ``Densely connected
  convolutional networks,'' in \emph{2017 {IEEE} Conference on Computer Vision
  and Pattern Recognition, {CVPR} 2017, Honolulu, HI, USA, July 21-26,
  2017}.\hskip 1em plus 0.5em minus 0.4em\relax {IEEE} Computer Society, 2017,
  pp. 2261--2269.

\bibitem{MobileNet}
M.~Sandler, A.~G. Howard, M.~Zhu, A.~Zhmoginov, and L.~Chen, ``Mobilenetv2:
  Inverted residuals and linear bottlenecks,'' in \emph{2018 {IEEE} Conference
  on Computer Vision and Pattern Recognition, {CVPR} 2018, Salt Lake City, UT,
  USA, June 18-22, 2018}.\hskip 1em plus 0.5em minus 0.4em\relax Computer
  Vision Foundation / {IEEE} Computer Society, 2018, pp. 4510--4520.

\bibitem{FGSM}
I.~J. Goodfellow, J.~Shlens, and C.~Szegedy, ``Explaining and harnessing
  adversarial examples,'' in \emph{3rd International Conference on Learning
  Representations, {ICLR} 2015, San Diego, CA, USA, May 7-9, 2015, Conference
  Track Proceedings}, 2015.

\bibitem{I-FGSM}
A.~Kurakin, I.~J. Goodfellow, and S.~Bengio, ``Adversarial examples in the
  physical world,'' in \emph{5th International Conference on Learning
  Representations, {ICLR} 2017, Toulon, France, April 24-26, 2017, Workshop
  Track Proceedings}.\hskip 1em plus 0.5em minus 0.4em\relax OpenReview.net,
  2017.

\bibitem{MI-FGSM}
Y.~Dong, F.~Liao, T.~Pang, H.~Su, J.~Zhu, X.~Hu, and J.~Li, ``Boosting
  adversarial attacks with momentum,'' in \emph{2018 {IEEE} Conference on
  Computer Vision and Pattern Recognition, {CVPR} 2018, Salt Lake City, UT,
  USA, June 18-22, 2018}.\hskip 1em plus 0.5em minus 0.4em\relax Computer
  Vision Foundation / {IEEE} Computer Society, 2018, pp. 9185--9193.

\bibitem{ZOO}
P.~Chen, H.~Zhang, Y.~Sharma, J.~Yi, and C.~Hsieh, ``{ZOO:} zeroth order
  optimization based black-box attacks to deep neural networks without training
  substitute models,'' in \emph{Proceedings of the 10th {ACM} Workshop on
  Artificial Intelligence and Security, AISec@CCS 2017, Dallas, TX, USA,
  November 3, 2017}.\hskip 1em plus 0.5em minus 0.4em\relax {ACM}, 2017, pp.
  15--26.

\bibitem{Square}
M.~Andriushchenko, F.~Croce, N.~Flammarion, and M.~Hein, ``Square attack: {A}
  query-efficient black-box adversarial attack via random search,'' in
  \emph{Computer Vision - {ECCV} 2020 - 16th European Conference, Glasgow, UK,
  August 23-28, 2020, Proceedings, Part {XXIII}}, ser. Lecture Notes in
  Computer Science, vol. 12368.\hskip 1em plus 0.5em minus 0.4em\relax
  Springer, 2020, pp. 484--501.

\bibitem{TRADES}
H.~Zhang, Y.~Yu, J.~Jiao, E.~P. Xing, L.~{El Ghaoui}, and M.~I. Jordan,
  ``Theoretically principled trade-off between robustness and accuracy,'' in
  \emph{Proceedings of the 36th International Conference on Machine Learning,
  {ICML} 2019, 9-15 June 2019, Long Beach, California, {USA}}, ser. Proceedings
  of Machine Learning Research, vol.~97.\hskip 1em plus 0.5em minus 0.4em\relax
  {PMLR}, 2019, pp. 7472--7482.

\bibitem{Fast-AT}
E.~Wong, L.~Rice, and J.~Z. Kolter, ``Fast is better than free: Revisiting
  adversarial training,'' in \emph{8th International Conference on Learning
  Representations, {ICLR} 2020, Addis Ababa, Ethiopia, April 26-30,
  2020}.\hskip 1em plus 0.5em minus 0.4em\relax OpenReview.net, 2020.

\bibitem{Bag-of-tricks-for-AT}
T.~Pang, X.~Yang, Y.~Dong, H.~Su, and J.~Zhu, ``Bag of tricks for adversarial
  training,'' in \emph{9th International Conference on Learning
  Representations, {ICLR} 2021, Virtual Event, Austria, May 3-7, 2021}.\hskip
  1em plus 0.5em minus 0.4em\relax OpenReview.net, 2021.

\bibitem{Madry-AT}
A.~Madry, A.~Makelov, L.~Schmidt, D.~Tsipras, and A.~Vladu, ``Towards deep
  learning models resistant to adversarial attacks,'' in \emph{6th
  International Conference on Learning Representations, {ICLR} 2018, Vancouver,
  BC, Canada, April 30 - May 3, 2018, Conference Track Proceedings}.\hskip 1em
  plus 0.5em minus 0.4em\relax OpenReview.net, 2018.

\bibitem{post-process-method}
S.~Chen, Z.~Huang, Q.~Tao, Y.~Wu, C.~Xie, and X.~Huang, ``Adversarial attack on
  attackers: Post-process to mitigate black-box score-based query attacks,''
  \emph{CoRR}, vol. abs/2205.12134, 2022.

\bibitem{Bandit}
A.~Ilyas, L.~Engstrom, and A.~Madry, ``Prior convictions: Black-box adversarial
  attacks with bandits and priors,'' in \emph{{ICLR} (Poster)}.\hskip 1em plus
  0.5em minus 0.4em\relax OpenReview.net, 2019.

\bibitem{SimBA}
C.~Guo, J.~R. Gardner, Y.~You, A.~G. Wilson, and K.~Q. Weinberger, ``Simple
  black-box adversarial attacks,'' in \emph{{ICML}}, ser. Proceedings of
  Machine Learning Research, vol.~97.\hskip 1em plus 0.5em minus 0.4em\relax
  {PMLR}, 2019, pp. 2484--2493.

\bibitem{SignHunter}
A.~Al{-}Dujaili and U.~O'Reilly, ``Sign bits are all you need for black-box
  attacks,'' in \emph{{ICLR}}.\hskip 1em plus 0.5em minus 0.4em\relax
  OpenReview.net, 2020.

\bibitem{NES}
A.~Ilyas, L.~Engstrom, A.~Athalye, and J.~Lin, ``Black-box adversarial attacks
  with limited queries and information,'' in \emph{{ICML}}, ser. Proceedings of
  Machine Learning Research, vol.~80.\hskip 1em plus 0.5em minus 0.4em\relax
  {PMLR}, 2018, pp. 2142--2151.

\bibitem{boundary-attack}
W.~Brendel, J.~Rauber, and M.~Bethge, ``Decision-based adversarial attacks:
  Reliable attacks against black-box machine learning models,'' in \emph{{ICLR}
  (Poster)}.\hskip 1em plus 0.5em minus 0.4em\relax OpenReview.net, 2018.

\bibitem{opt-attack}
M.~Cheng, T.~Le, P.~Chen, H.~Zhang, J.~Yi, and C.~Hsieh, ``Query-efficient
  hard-label black-box attack: An optimization-based approach,'' in
  \emph{{ICLR} (Poster)}.\hskip 1em plus 0.5em minus 0.4em\relax
  OpenReview.net, 2019.

\bibitem{HopSkipJumpAttack}
J.~Chen, M.~I. Jordan, and M.~J. Wainwright, ``Hopskipjumpattack: {A}
  query-efficient decision-based attack,'' in \emph{{IEEE} Symposium on
  Security and Privacy}.\hskip 1em plus 0.5em minus 0.4em\relax {IEEE}, 2020,
  pp. 1277--1294.

\bibitem{Sign-opt}
M.~Cheng, S.~Singh, P.~H. Chen, P.~Chen, S.~Liu, and C.~Hsieh, ``Sign-opt: {A}
  query-efficient hard-label adversarial attack,'' in \emph{{ICLR}}.\hskip 1em
  plus 0.5em minus 0.4em\relax OpenReview.net, 2020.

\bibitem{P-RGF}
S.~Cheng, Y.~Dong, T.~Pang, H.~Su, and J.~Zhu, ``Improving black-box
  adversarial attacks with a transfer-based prior,'' in \emph{Advances in
  Neural Information Processing Systems 32: Annual Conference on Neural
  Information Processing Systems 2019, NeurIPS 2019, December 8-14, 2019,
  Vancouver, BC, Canada}, 2019, pp. 10\,932--10\,942.

\bibitem{TREMBA}
Z.~Huang and T.~Zhang, ``Black-box adversarial attack with transferable
  model-based embedding,'' in \emph{{ICLR}}.\hskip 1em plus 0.5em minus
  0.4em\relax OpenReview.net, 2020.

\bibitem{LeBA}
J.~Yang, Y.~Jiang, X.~Huang, B.~Ni, and C.~Zhao, ``Learning black-box attackers
  with transferable priors and query feedback,'' in \emph{Advances in Neural
  Information Processing Systems 33: Annual Conference on Neural Information
  Processing Systems 2020, NeurIPS 2020, December 6-12, 2020, virtual}, 2020.

\bibitem{ODS}
Y.~Tashiro, Y.~Song, and S.~Ermon, ``Diversity can be transferred: Output
  diversification for white- and black-box attacks,'' in \emph{Advances in
  Neural Information Processing Systems 33: Annual Conference on Neural
  Information Processing Systems 2020, NeurIPS 2020, December 6-12, 2020,
  virtual}, 2020.

\bibitem{GFCS}
N.~A. Lord, R.~M{\"{u}}ller, and L.~Bertinetto, ``Attacking deep networks with
  surrogate-based adversarial black-box methods is easy,'' in
  \emph{{ICLR}}.\hskip 1em plus 0.5em minus 0.4em\relax OpenReview.net, 2022.

\bibitem{DI-FGSM}
C.~Xie, Z.~Zhang, Y.~Zhou, S.~Bai, J.~Wang, Z.~Ren, and A.~L. Yuille,
  ``Improving transferability of adversarial examples with input diversity,''
  in \emph{{IEEE} Conference on Computer Vision and Pattern Recognition, {CVPR}
  2019, Long Beach, CA, USA, June 16-20, 2019}.\hskip 1em plus 0.5em minus
  0.4em\relax Computer Vision Foundation / {IEEE}, 2019, pp. 2730--2739.

\bibitem{SI-NI-FGSM}
J.~Lin, C.~Song, K.~He, L.~Wang, and J.~E. Hopcroft, ``Nesterov accelerated
  gradient and scale invariance for adversarial attacks,'' in \emph{8th
  International Conference on Learning Representations, {ICLR} 2020, Addis
  Ababa, Ethiopia, April 26-30, 2020}.\hskip 1em plus 0.5em minus 0.4em\relax
  OpenReview.net, 2020.

\bibitem{VMI-FGSM}
X.~Wang and K.~He, ``Enhancing the transferability of adversarial attacks
  through variance tuning,'' in \emph{{IEEE} Conference on Computer Vision and
  Pattern Recognition, {CVPR} 2021, virtual, June 19-25, 2021}.\hskip 1em plus
  0.5em minus 0.4em\relax Computer Vision Foundation / {IEEE}, 2021, pp.
  1924--1933.

\bibitem{gradient-alignment}
A.~Demontis, M.~Melis, M.~Pintor, M.~Jagielski, B.~Biggio, A.~Oprea,
  C.~Nita{-}Rotaru, and F.~Roli, ``Why do adversarial attacks transfer?
  explaining transferability of evasion and poisoning attacks,'' in
  \emph{{USENIX} Security Symposium}.\hskip 1em plus 0.5em minus 0.4em\relax
  {USENIX} Association, 2019, pp. 321--338.

\bibitem{SGM}
D.~Wu, Y.~Wang, S.~Xia, J.~Bailey, and X.~Ma, ``Skip connections matter: On the
  transferability of adversarial examples generated with resnets,'' in
  \emph{8th International Conference on Learning Representations, {ICLR} 2020,
  Addis Ababa, Ethiopia, April 26-30, 2020}.\hskip 1em plus 0.5em minus
  0.4em\relax OpenReview.net, 2020.

\bibitem{MoreBayesian}
Q.~Li, Y.~Guo, W.~Zuo, and H.~Chen, ``Making substitute models more bayesian
  can enhance transferability of adversarial examples,'' \emph{CoRR}, vol.
  abs/2302.05086, 2023.

\bibitem{Dual-Path-Distillation}
Y.~Zhang, Y.~Li, T.~Liu, and X.~Tian, ``Dual-path distillation: {A} unified
  framework to improve black-box attacks,'' in \emph{Proceedings of the 37th
  International Conference on Machine Learning, {ICML} 2020, 13-18 July 2020,
  Virtual Event}, ser. Proceedings of Machine Learning Research, vol.
  119.\hskip 1em plus 0.5em minus 0.4em\relax {PMLR}, 2020, pp.
  11\,163--11\,172.

\bibitem{DTA}
X.~Yang, J.~Lin, H.~Zhang, X.~Yang, and P.~Zhao, ``Improving the
  transferability of adversarial examples via direction tuning,'' \emph{CoRR},
  vol. abs/2303.15109, 2023.

\bibitem{ImageNet}
O.~Russakovsky, J.~Deng, H.~Su, J.~Krause, S.~Satheesh, S.~Ma, Z.~Huang,
  A.~Karpathy, A.~Khosla, M.~S. Bernstein, A.~C. Berg, and L.~Fei{-}Fei,
  ``Imagenet large scale visual recognition challenge,'' \emph{Int. J. Comput.
  Vis.}, vol. 115, no.~3, pp. 211--252, 2015.

\bibitem{TI-FGSM}
Y.~Dong, T.~Pang, H.~Su, and J.~Zhu, ``Evading defenses to transferable
  adversarial examples by translation-invariant attacks,'' in
  \emph{{CVPR}}.\hskip 1em plus 0.5em minus 0.4em\relax Computer Vision
  Foundation / {IEEE}, 2019, pp. 4312--4321.

\bibitem{CW}
N.~Carlini and D.~A. Wagner, ``Towards evaluating the robustness of neural
  networks,'' in \emph{{IEEE} Symposium on Security and Privacy}.\hskip 1em
  plus 0.5em minus 0.4em\relax {IEEE} Computer Society, 2017, pp. 39--57.

\bibitem{DeepFool}
S.~Moosavi{-}Dezfooli, A.~Fawzi, and P.~Frossard, ``Deepfool: {A} simple and
  accurate method to fool deep neural networks,'' in \emph{{CVPR}}.\hskip 1em
  plus 0.5em minus 0.4em\relax {IEEE} Computer Society, 2016, pp. 2574--2582.

\bibitem{JSMA}
N.~Papernot, P.~D. McDaniel, S.~Jha, M.~Fredrikson, Z.~B. Celik, and A.~Swami,
  ``The limitations of deep learning in adversarial settings,'' in
  \emph{EuroS{\&}P}.\hskip 1em plus 0.5em minus 0.4em\relax {IEEE}, 2016, pp.
  372--387.

\bibitem{Pixel-attack}
J.~Su, D.~V. Vargas, and K.~Sakurai, ``One pixel attack for fooling deep neural
  networks,'' \emph{{IEEE} Trans. Evol. Comput.}, vol.~23, no.~5, pp. 828--841,
  2019.

\bibitem{AutoAttack}
F.~Croce and M.~Hein, ``Reliable evaluation of adversarial robustness with an
  ensemble of diverse parameter-free attacks,'' in \emph{{ICML}}, ser.
  Proceedings of Machine Learning Research, vol. 119.\hskip 1em plus 0.5em
  minus 0.4em\relax {PMLR}, 2020, pp. 2206--2216.

\bibitem{Free-AT}
A.~Shafahi, M.~Najibi, A.~Ghiasi, Z.~Xu, J.~P. Dickerson, C.~Studer, L.~S.
  Davis, G.~Taylor, and T.~Goldstein, ``Adversarial training for free!'' in
  \emph{NeurIPS}, 2019, pp. 3353--3364.

\bibitem{RCE}
C.~Zhang, P.~Benz, A.~Karjauv, J.~Cho, K.~Zhang, and I.~S. Kweon,
  ``Investigating top-k white-box and transferable black-box attack,''
  \emph{CoRR}, vol. abs/2204.00089, 2022.

\bibitem{RAP}
Z.~Qin, Y.~Fan, Y.~Liu, L.~Shen, Y.~Zhang, J.~Wang, and B.~Wu, ``Boosting the
  transferability of adversarial attacks with reverse adversarial
  perturbation,'' \emph{CoRR}, vol. abs/2210.05968, 2022.

\bibitem{FIA}
Z.~Wang, H.~Guo, Z.~Zhang, W.~Liu, Z.~Qin, and K.~Ren, ``Feature
  importance-aware transferable adversarial attacks,'' in \emph{2021 {IEEE/CVF}
  International Conference on Computer Vision, {ICCV} 2021, Montreal, QC,
  Canada, October 10-17, 2021}.\hskip 1em plus 0.5em minus 0.4em\relax {IEEE},
  2021, pp. 7619--7628.

\bibitem{NAA}
J.~Zhang, W.~Wu, J.~Huang, Y.~Huang, W.~Wang, Y.~Su, and M.~R. Lyu, ``Improving
  adversarial transferability via neuron attribution-based attacks,''
  \emph{CoRR}, vol. abs/2204.00008, 2022.

\bibitem{ATA}
W.~Wu, Y.~Su, X.~Chen, S.~Zhao, I.~King, M.~R. Lyu, and Y.~Tai, ``Boosting the
  transferability of adversarial samples via attention,'' in \emph{2020
  {IEEE/CVF} Conference on Computer Vision and Pattern Recognition, {CVPR}
  2020, Seattle, WA, USA, June 13-19, 2020}.\hskip 1em plus 0.5em minus
  0.4em\relax Computer Vision Foundation / {IEEE}, 2020, pp. 1158--1167.

\bibitem{AA}
N.~Inkawhich, W.~Wen, H.~H. Li, and Y.~Chen, ``Feature space perturbations
  yield more transferable adversarial examples,'' in \emph{{IEEE} Conference on
  Computer Vision and Pattern Recognition, {CVPR} 2019, Long Beach, CA, USA,
  June 16-20, 2019}.\hskip 1em plus 0.5em minus 0.4em\relax Computer Vision
  Foundation / {IEEE}, 2019, pp. 7066--7074.

\bibitem{FDA+xent}
N.~Inkawhich, K.~J. Liang, B.~Wang, M.~Inkawhich, L.~Carin, and Y.~Chen,
  ``Perturbing across the feature hierarchy to improve standard and strict
  blackbox attack transferability,'' in \emph{Advances in Neural Information
  Processing Systems 33: Annual Conference on Neural Information Processing
  Systems 2020, NeurIPS 2020, December 6-12, 2020, virtual}, 2020.

\bibitem{ILA}
Q.~Huang, I.~Katsman, Z.~Gu, H.~He, S.~J. Belongie, and S.~Lim, ``Enhancing
  adversarial example transferability with an intermediate level attack,'' in
  \emph{2019 {IEEE/CVF} International Conference on Computer Vision, {ICCV}
  2019, Seoul, Korea (South), October 27 - November 2, 2019}.\hskip 1em plus
  0.5em minus 0.4em\relax {IEEE}, 2019, pp. 4732--4741.

\bibitem{FDA}
A.~Ganeshan, V.~B. S., and V.~B. Radhakrishnan, ``{FDA:} feature disruptive
  attack,'' in \emph{2019 {IEEE/CVF} International Conference on Computer
  Vision, {ICCV} 2019, Seoul, Korea (South), October 27 - November 2,
  2019}.\hskip 1em plus 0.5em minus 0.4em\relax {IEEE}, 2019, pp. 8068--8078.

\bibitem{ILA++}
Q.~Li, Y.~Guo, and H.~Chen, ``Yet another intermediate-level attack,'' in
  \emph{Computer Vision - {ECCV} 2020 - 16th European Conference, Glasgow, UK,
  August 23-28, 2020, Proceedings, Part {XVI}}, ser. Lecture Notes in Computer
  Science, vol. 12361.\hskip 1em plus 0.5em minus 0.4em\relax Springer, 2020,
  pp. 241--257.

\bibitem{LinBP}
Y.~Guo, Q.~Li, and H.~Chen, ``Backpropagating linearly improves transferability
  of adversarial examples,'' in \emph{Advances in Neural Information Processing
  Systems 33: Annual Conference on Neural Information Processing Systems 2020,
  NeurIPS 2020, December 6-12, 2020, virtual}, 2020.

\bibitem{LGV}
M.~Gubri, M.~Cordy, M.~Papadakis, Y.~L. Traon, and K.~Sen, ``{LGV:} boosting
  adversarial example transferability from large geometric vicinity,'' in
  \emph{{ECCV} {(4)}}, ser. Lecture Notes in Computer Science, vol.
  13664.\hskip 1em plus 0.5em minus 0.4em\relax Springer, 2022, pp. 603--618.

\bibitem{Admix}
X.~Wang, X.~He, J.~Wang, and K.~He, ``Admix: Enhancing the transferability of
  adversarial attacks,'' in \emph{2021 {IEEE/CVF} International Conference on
  Computer Vision, {ICCV} 2021, Montreal, QC, Canada, October 10-17,
  2021}.\hskip 1em plus 0.5em minus 0.4em\relax {IEEE}, 2021, pp.
  16\,138--16\,147.

\bibitem{Latent-HSJA}
D.~Na, S.~Ji, and J.~Kim, ``Unrestricted black-box adversarial attack using
  {GAN} with limited queries,'' in \emph{{ECCV} Workshops {(1)}}, ser. Lecture
  Notes in Computer Science, vol. 13801.\hskip 1em plus 0.5em minus 0.4em\relax
  Springer, 2022, pp. 467--482.

\bibitem{NCF}
S.~Yuan, Q.~Zhang, L.~Gao, Y.~Cheng, and J.~Song, ``Natural color fool: Towards
  boosting black-box unrestricted attacks,'' \emph{CoRR}, vol. abs/2210.02041,
  2022.

\bibitem{GARSDC}
S.~Liang, L.~Li, Y.~Fan, X.~Jia, J.~Li, B.~Wu, and X.~Cao, ``A large-scale
  multiple-objective method for black-box attack against object detection,'' in
  \emph{{ECCV} {(4)}}, ser. Lecture Notes in Computer Science, vol.
  13664.\hskip 1em plus 0.5em minus 0.4em\relax Springer, 2022, pp. 619--636.

\bibitem{Feature-level-transformation}
G.~Chen, Z.~Zhao, F.~Song, S.~Chen, L.~Fan, F.~Wang, and J.~Wang, ``Towards
  understanding and mitigating audio adversarial examples for speaker
  recognition,'' \emph{IEEE Transactions on Dependable and Secure Computing},
  pp. 1--17, 2022.

\bibitem{Text-Fooler}
D.~Jin, Z.~Jin, J.~T. Zhou, and P.~Szolovits, ``Is {BERT} really robust? {A}
  strong baseline for natural language attack on text classification and
  entailment,'' in \emph{{AAAI}}.\hskip 1em plus 0.5em minus 0.4em\relax {AAAI}
  Press, 2020, pp. 8018--8025.

\bibitem{Bert-attack}
L.~Li, R.~Ma, Q.~Guo, X.~Xue, and X.~Qiu, ``{BERT-ATTACK:} adversarial attack
  against {BERT} using {BERT},'' in \emph{{EMNLP} {(1)}}.\hskip 1em plus 0.5em
  minus 0.4em\relax Association for Computational Linguistics, 2020, pp.
  6193--6202.

\bibitem{POC}
G.~Apruzzese and V.~S. Subrahmanian, ``Mitigating adversarial gray-box attacks
  against phishing detectors,'' \emph{IEEE Transactions on Dependable and
  Secure Computing}, pp. 1--19, 2022.

\bibitem{GVSA}
Q.~Song, R.~Tan, C.~Ren, Y.~Xu, Y.~Lou, J.~Wang, and H.~B. Gooi, ``On
  credibility of adversarial examples against learning-based grid voltage
  stability assessment,'' \emph{IEEE Transactions on Dependable and Secure
  Computing}, pp. 1--14, 2022.

\bibitem{UniAP}
P.~Cheng, Y.~Wu, Y.~Hong, Z.~Ba, F.~Lin, L.~Lu, and K.~Ren, ``Uniap: Protecting
  speech privacy with non-targeted universal adversarial perturbations,''
  \emph{IEEE Transactions on Dependable and Secure Computing}, pp. 1--16, 2023.

\bibitem{YOPO}
D.~Zhang, T.~Zhang, Y.~Lu, Z.~Zhu, and B.~Dong, ``You only propagate once:
  Accelerating adversarial training via maximal principle,'' in \emph{NeurIPS},
  2019, pp. 227--238.

\bibitem{ATTA}
H.~Zheng, Z.~Zhang, J.~Gu, H.~Lee, and A.~Prakash, ``Efficient adversarial
  training with transferable adversarial examples,'' in \emph{{CVPR}}.\hskip
  1em plus 0.5em minus 0.4em\relax Computer Vision Foundation / {IEEE}, 2020,
  pp. 1178--1187.

\bibitem{FAT}
J.~Zhang, X.~Xu, B.~Han, G.~Niu, L.~Cui, M.~Sugiyama, and M.~S. Kankanhalli,
  ``Attacks which do not kill training make adversarial learning stronger,'' in
  \emph{{ICML}}, ser. Proceedings of Machine Learning Research, vol. 119.\hskip
  1em plus 0.5em minus 0.4em\relax {PMLR}, 2020, pp. 11\,278--11\,287.

\bibitem{LAS-AT}
X.~Jia, Y.~Zhang, B.~Wu, K.~Ma, J.~Wang, and X.~Cao, ``{LAS-AT:} adversarial
  training with learnable attack strategy,'' in \emph{{CVPR}}.\hskip 1em plus
  0.5em minus 0.4em\relax {IEEE}, 2022, pp. 13\,388--13\,398.

\bibitem{Fixing-data}
S.~Rebuffi, S.~Gowal, D.~A. Calian, F.~Stimberg, O.~Wiles, and T.~A. Mann,
  ``Fixing data augmentation to improve adversarial robustness,'' \emph{CoRR},
  vol. abs/2103.01946, 2021.

\bibitem{AdvAug}
D.~A. Calian, F.~Stimberg, O.~Wiles, S.~Rebuffi, A.~Gy{\"{o}}rgy, T.~A. Mann,
  and S.~Gowal, ``Defending against image corruptions through adversarial
  augmentations,'' in \emph{{ICLR}}.\hskip 1em plus 0.5em minus 0.4em\relax
  OpenReview.net, 2022.

\bibitem{AugMix}
D.~Hendrycks, N.~Mu, E.~D. Cubuk, B.~Zoph, J.~Gilmer, and B.~Lakshminarayanan,
  ``Augmix: {A} simple data processing method to improve robustness and
  uncertainty,'' in \emph{{ICLR}}.\hskip 1em plus 0.5em minus 0.4em\relax
  OpenReview.net, 2020.

\bibitem{RandAugment}
E.~D. Cubuk, B.~Zoph, J.~Shlens, and Q.~Le, ``Randaugment: Practical automated
  data augmentation with a reduced search space,'' in \emph{NeurIPS}, 2020.

\bibitem{NoisyMix}
N.~B. Erichson, S.~H. Lim, F.~Utrera, W.~Xu, Z.~Cao, and M.~W. Mahoney,
  ``Noisymix: Boosting robustness by combining data augmentations, stability
  training, and noise injections,'' \emph{CoRR}, vol. abs/2202.01263, 2022.

\bibitem{Proxy-distributions}
V.~Sehwag, S.~Mahloujifar, T.~Handina, S.~Dai, C.~Xiang, M.~Chiang, and
  P.~Mittal, ``Robust learning meets generative models: Can proxy distributions
  improve adversarial robustness?'' in \emph{{ICLR}}.\hskip 1em plus 0.5em
  minus 0.4em\relax OpenReview.net, 2022.

\bibitem{CARDs}
J.~Diffenderfer, B.~R. Bartoldson, S.~Chaganti, J.~Zhang, and B.~Kailkhura, ``A
  winning hand: Compressing deep networks can improve out-of-distribution
  robustness,'' in \emph{NeurIPS}, 2021, pp. 664--676.

\bibitem{RobustWRN}
H.~Huang, Y.~Wang, S.~M. Erfani, Q.~Gu, J.~Bailey, and X.~Ma, ``Exploring
  architectural ingredients of adversarially robust deep neural networks,'' in
  \emph{NeurIPS}, 2021, pp. 5545--5559.

\bibitem{RobustResNets}
S.~Huang, Z.~Lu, K.~Deb, and V.~N. Boddeti, ``Revisiting residual networks for
  adversarial robustness: An architectural perspective,'' \emph{CoRR}, vol.
  abs/2212.11005, 2022.

\bibitem{Feature-Squeezing}
W.~Xu, D.~Evans, and Y.~Qi, ``Feature squeezing: Detecting adversarial examples
  in deep neural networks,'' in \emph{{NDSS}}.\hskip 1em plus 0.5em minus
  0.4em\relax The Internet Society, 2018.

\bibitem{MPD}
G.~Vacanti and A.~V. Looveren, ``Adversarial detection and correction by
  matching prediction distributions,'' \emph{CoRR}, vol. abs/2002.09364, 2020.

\bibitem{ADDITION}
Y.~Wang, X.~Li, L.~Yang, J.~Ma, and H.~Li, ``Addition: Detecting adversarial
  examples with image-dependent noise reduction,'' \emph{IEEE Transactions on
  Dependable and Secure Computing}, pp. 1--16, 2023.

\bibitem{LID}
X.~Ma, B.~Li, Y.~Wang, S.~M. Erfani, S.~N.~R. Wijewickrema, G.~Schoenebeck,
  D.~Song, M.~E. Houle, and J.~Bailey, ``Characterizing adversarial subspaces
  using local intrinsic dimensionality,'' in \emph{{ICLR}}.\hskip 1em plus
  0.5em minus 0.4em\relax OpenReview.net, 2018.

\bibitem{MD}
K.~Lee, K.~Lee, H.~Lee, and J.~Shin, ``A simple unified framework for detecting
  out-of-distribution samples and adversarial attacks,'' in \emph{NeurIPS},
  2018, pp. 7167--7177.

\bibitem{KDBU}
R.~Feinman, R.~R. Curtin, S.~Shintre, and A.~B. Gardner, ``Detecting
  adversarial samples from artifacts,'' \emph{CoRR}, vol. abs/1703.00410, 2017.

\bibitem{NIC}
S.~Ma, Y.~Liu, G.~Tao, W.~Lee, and X.~Zhang, ``{NIC:} detecting adversarial
  samples with neural network invariant checking,'' in \emph{{NDSS}}.\hskip 1em
  plus 0.5em minus 0.4em\relax The Internet Society, 2019.

\bibitem{Adversarial-cone}
K.~Roth, Y.~Kilcher, and T.~Hofmann, ``The odds are odd: {A} statistical test
  for detecting adversarial examples,'' in \emph{{ICML}}, ser. Proceedings of
  Machine Learning Research, vol.~97.\hskip 1em plus 0.5em minus 0.4em\relax
  {PMLR}, 2019, pp. 5498--5507.

\bibitem{LiBRe}
Z.~Deng, X.~Yang, S.~Xu, H.~Su, and J.~Zhu, ``Libre: {A} practical bayesian
  approach to adversarial detection,'' in \emph{{CVPR}}.\hskip 1em plus 0.5em
  minus 0.4em\relax Computer Vision Foundation / {IEEE}, 2021, pp. 972--982.

\bibitem{JTLA}
J.~Raghuram, V.~Chandrasekaran, S.~Jha, and S.~Banerjee, ``A general framework
  for detecting anomalous inputs to {DNN} classifiers,'' in \emph{{ICML}}, ser.
  Proceedings of Machine Learning Research, vol. 139.\hskip 1em plus 0.5em
  minus 0.4em\relax {PMLR}, 2021, pp. 8764--8775.

\bibitem{Provably-robust-classification}
F.~Sheikholeslami, A.~Lotfi, and J.~Z. Kolter, ``Provably robust classification
  of adversarial examples with detection,'' in \emph{{ICLR}}.\hskip 1em plus
  0.5em minus 0.4em\relax OpenReview.net, 2021.

\bibitem{Hardness-Reduction}
F.~Tram{\`{e}}r, ``Detecting adversarial examples is (nearly) as hard as
  classifying them,'' in \emph{{ICML}}, ser. Proceedings of Machine Learning
  Research, vol. 162.\hskip 1em plus 0.5em minus 0.4em\relax {PMLR}, 2022, pp.
  21\,692--21\,702.

\end{thebibliography}

\end{document}